\pdfoutput=1
\documentclass[letterpaper]{article} 
\usepackage{aaai25}  
\usepackage{times}  
\usepackage{helvet}  
\usepackage{courier}  
\usepackage[hyphens]{url}  
\usepackage{graphicx} 
\usepackage{natbib}  
\usepackage{caption} 
\usepackage{algorithm}
\usepackage{algorithmic}
\usepackage{algorithm}
\usepackage{algorithmic}
\usepackage{amsmath}
\usepackage{dsfont}
\usepackage{makecell}
\usepackage{amssymb}
\usepackage{booktabs} 
\usepackage{tabularx}
\usepackage{newfloat}
\usepackage{listings}
\DeclareCaptionStyle{ruled}{labelfont=normalfont,labelsep=colon,strut=off} 
\floatstyle{ruled}
\newfloat{listing}{tb}{lst}{}
\floatname{listing}{Listing}
\usepackage{xcolor} 

\title{Multi-StyleGS: Stylizing Gaussian Splatting 
 with Multiple Styles}
\author{
    Yangkai Lin\textsuperscript{\rm 1}, Jiabao Lei\textsuperscript{\rm 2}, Kui jia\textsuperscript{\rm 2}\thanks{Corresponding author.}
}
\affiliations{
    \textsuperscript{\rm 1}South China University of Technology\\
    \textsuperscript{\rm 2}School of Data Science, The Chinese University of Hong Kong, Shenzhen \\
    202210182091@mail.scut.edu.cn, jiabaolei@link.cuhk.edu.cn, 
    kuijia@cuhk.edu.cn
}

\usepackage{bibentry}

\begin{document}

\makeatletter
\let\@oldmaketitle\@maketitle
\renewcommand{\@maketitle}{\@oldmaketitle
  \includegraphics[width=\linewidth]{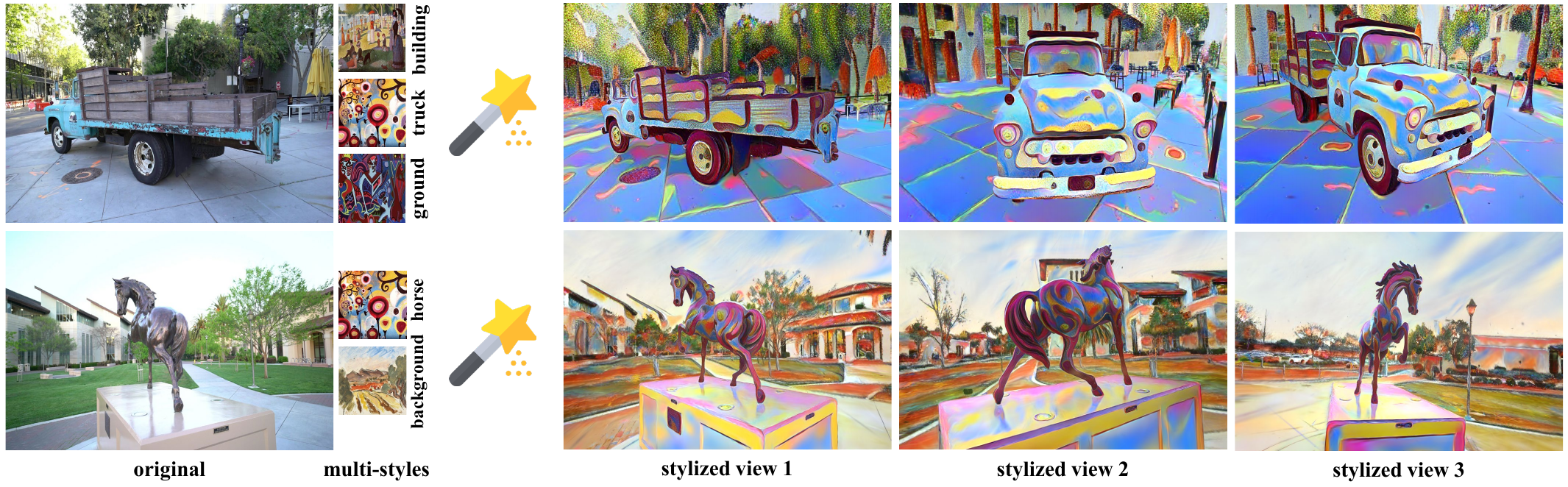}
  \captionof{figure}{With a set of multi-view images of a 3D scene and several specified style images, our method can transfer artistic styles to the 3D scene, creating high-quality stylized images of novel views with consistency.}\label{teaser_fig}
  \bigskip
    }
\makeatother

\maketitle

  

\begin{abstract}
  In recent years, there has been a growing demand to stylize a given 3D scene to align with the artistic style of reference images for creative purposes. While 3D Gaussian Splatting (GS) has emerged as a promising and efficient method for realistic 3D scene modeling, there remains a challenge in adapting it to stylize 3D GS to match with multiple styles through automatic local style transfer or manual designation, while maintaining memory efficiency for stylization training. In this paper, we introduce a novel 3D GS stylization solution termed Multi-StyleGS to tackle these challenges. In particular, we employ a bipartite matching mechanism to automatically identify correspondences between the style images and the local regions of the rendered images. To facilitate local style transfer, we introduce a novel semantic style loss function that employs a segmentation network to apply distinct styles to various objects of the scene and propose a local-global feature matching to enhance the multi-view consistency. Furthermore, this technique can achieve memory-efficient training, more texture details and better color match. To better assign a robust semantic label to each Gaussian, we propose several techniques to regularize the segmentation network. As demonstrated by our comprehensive experiments, our approach outperforms existing ones in producing plausible stylization results and offering flexible editing. code: \url{https://github.com/SCUTykLin/Multi-StyleGS.git}
\end{abstract}

%

\section{Introduction}
Artistic creation has attracted considerable attention, with aesthetic 3D content creation being one of the urgent demands in recent years. Stylizing an already acquired 3D scene is the primary approach to obtain artistic 3D content.
In this paper, our focus is on the task of 3D scene stylization, where we aim to transfer reference styles specified by multiple style images to the 3D scene.

%
Previous work on 3D stylization~\cite{stylizednerf, nerf-art, lsnerf, geometryTransfer} has predominantly utilized the Neural Radiance Field (NeRF) as a scene representation \cite{nerf}. While NeRF is compact and capable of achieving photorealistic rendering results, it is limited to implicit editing and faces significant performance challenges due to the utilization of a heavy and high-dimensional Multi-Layer Perceptron (MLP) network for scene representation. Balancing computational time and result quality requires a delicate trade-off. Despite some advancements \cite{instantngp, Kilonerf, DVGO, plenoxels, tensorf, TriMipRF, ZipNeRF} aimed at mitigating the performance issues of NeRF in practical applications, these challenges still persist.
%
Recently, a significant portion of the work on 3D stylization has concentrated on global stylization~\cite{ARF, snerf, styleRF, HyperNetwork}, where the same style pattern is applied uniformly to all parts of the 3D content. However, this approach can be suboptimal as not all regions should be treated equally, limiting flexibility and editability. Another portion of the work focus on the local stylization~\cite{ConRF, ref-npr}. However, they can only stylize simple scenes~\cite{LLFF} and struggle to ensure multi-view consistency. Furthermore, techniques that employ Gaussian Splatting (GS) \cite{gs} frequently encounter memory bottleneck issues, impeding the progress for further applications.

To address these challenges, we introduce a novel 3D stylization solution called Multi-StyleGS. This method is designed to deliver flexible and efficient image-based stylization of 3D scenes by empowering explicit local editing.

Specifically, we choose GS~\cite{gs} as our base representation, for its real-time rendering performance and explicit characteristic. While promising, we have noted a substantial increase in memory usage and the emergence of multi-view inconsistency in feature matching. To address these challenges, we introduce a novel semantic style loss to mitigate the problem of excessive memory consumption and multi-view inconsistency. Furthermore, to enable local stylization on semantic regions, we introduce an additional semantic feature for each GS, and update them during optimization.
This enhancement facilitates automatic local style transfer for region correspondences between multiple style images and the 3D scene.
%

Technically, we perform local style transfer as an additional post-processing step after capturing the original geometry and appearance of the 3D scenes using GS.
During the stylization process, we solely optimize the appearance of Gaussians. In addition to reconstruct 3D scene, we introduce an extra segmentation attribute that divides the Gaussians of the scene into multiple parts and paired with multiple style images through an effective bipartite matching mechanism~\cite{deepphoto} to automatically establish local region correspondences based on their feature similarity or manual designation. Subsequently, a novel multi-style loss is applied to guarantee local editability. Additionaly, we observed a multi-view inconsistency issue. Inspired by \cite{Probing3D}, we utilize DINOv2\cite{dinov2} to extract global features and introduce local-global matching for enhanced multi-view consistency with our novel multi-style loss, improving consistency, texture details, and color accuracy.
To address the potential issue of segmentation error due to the high degree of freedom of Gaussians, we introduce Gaussian smoothing regularization to alleviate this problem. Additionally, in order to mitigate the semantic ambiguity problem, we develop a technique called semantic importance filtering, which leverages semantic labels to effectively eliminate those Gaussians exhibiting semantic uncertainty; a negative entropy regularization term is also applied to each Gaussian to enforce semantic clarity. Our segmentation approach leverages SAM's \cite{SAM} capability, which we apply to the 3D scene to enhance multi-view consistency.


Our solution is able to handle styles from one single image or multiple images.
Extensive experiments conducted on various datasets \cite{tankandtemples, LLFF} substantiate the efficacy of our method in generating high-quality, locally matched stylized images in real-time. 
To summarize, our main contributions are:
\begin{itemize}
    \vspace*{-0.3em}
    \item A novel GS-based approach for local stylization of 3D scenes, facilitating the transfer of multiple artistic styles from one or several 2D images to 3D scenes.
    \item A new style loss that adopts bipartite matching assignment between multiple style image regions and GS points to enable (automatic) local style transfer. 
    \item A local-global feature matching solution to improve multi-view consistency. 
    \item Several regularization terms for removing noisy Gaussians and accuracy segmentation.
\end{itemize}

\section{Related Works}

\subsection{Overview of 3D Style Transfer}
Conventional approaches to stylize 3D scenes use explicit representations like point clouds~\cite{stylizenovelviews, 3dphotostylization} or meshes~\cite{3dphotostylization, Text2mesh, meshrenderer, Stylemesh}.
These approaches, however, is error-prone and may fail to capture geometry and texture details. 
NeRF~\cite{nerf} encodes a 3D scene using a neural network, making it a more suitable representation for downstream stylization tasks compared to explicit ones.
A common approach to stylizing a NeRF is to optimize and constrain its rendered images to a specific style using content loss and style loss.
snerf, arf and ins~\cite{snerf, ARF, unifiedimplicit} follow this line and optimizes neural networks using style loss. snerf renders blurry results due to refine geometry without supervision, and arf fixs the geometry branch and proposes nearest-neighbor feature matching loss to capture details. ins decouples NeRF to allow for separately encoding of representations.

Their results do not support diverse stylization results and typically stylize only the foreground of the scenes. 
HyperNet \cite{stylizing3Dscene} uses a hypernetwork to predict the weights of MLP to speed up stylization. LsNeRF \cite{lsnerf} introduces a region-matching style loss designed to enhance local stylization of the 3D scenes. Yet, this method faces limitations as it cannot concurrently assimilate styles from multiple images into a single 3D scene. Moreover, it is unable to maintain multi-view stylistic consistency. Our work introduces a matching mechanism~\cite{lsnerf} to establish region correspondences and a novel style loss to support local style transfer. Besides, thanks to the use of explicit representation of GS~\cite{gs}, we can nicely stylize the background as well.

\begin{figure*}[tb]
  \centering
  \includegraphics[width=1.0\textwidth]{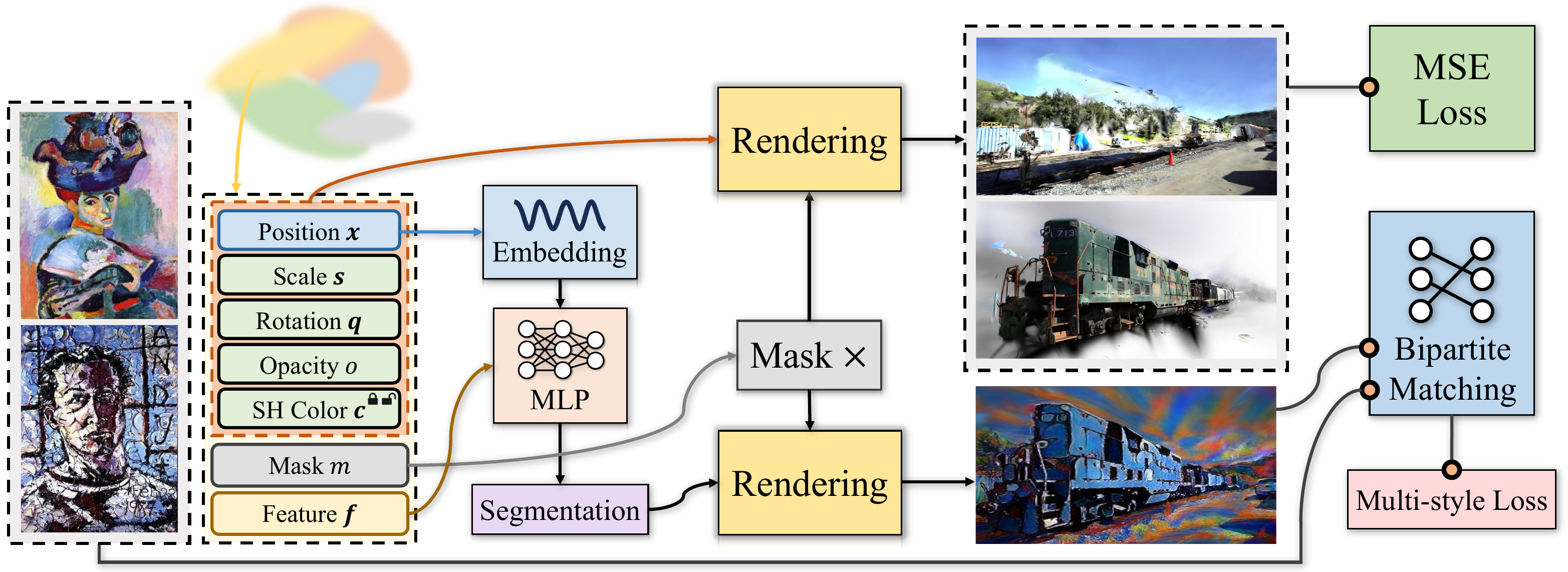}
  \caption{
  Overview of our pipeline. It firstly reconstructs a GS model from multiple training images, and then stylize the scene using
bipartite matching with multiple styles. Upon completion, it can produce consistent free-viewpoint stylized renderings.
  \vspace{-0.5cm}
  }
  \label{fig:style_stage}
\end{figure*}

\vspace{-0.2cm}
\subsection{Memory-efficient 3D Style Transfer}
However, such methods are very memory-inefficient in practice. ARF~\cite{ARF} propose a deferred back-propagation method to enable optimization of memory-intensive NeRF. StyleRF~\cite{styleRF} proposes a deferred style transformation of 2D feature maps to greatly reduces memory footprint. These approaches are all developed based on NeRF~\cite{nerf} and are not applicable to GS~\cite{gs}. Our novel semantic style loss can achieve memory-efficient training, which enables efficient training on a single RTX 3090.

\vspace{-0.2cm}
\subsection{3D Local Stylization}
Another line of work \cite{lsnerf, ref-npr, ConRF} investigates local stylization methods, which allow for diverse styles on local regions. However, most of the work can only stylize relatively simple scenes and cannot ensure multi-view consistency. Our method proposes a local-global matching to tackle this issue and conducts 
extensive experiments on various datasets\cite{LLFF, tankandtemples}.

\section{Preliminary of Gaussian Splatting}
Gaussian Splatting~(GS)~\cite{gs} represents a 3D scene with a set of 3D Gaussians. Each Gaussian consists of a center location $\boldsymbol{\mu}\in\mathbb{R}^3$, a covariance matrix $\mathbf{\Sigma}\in\mathbb{R}^{3\times3}$. The covariance matrix $\mathbf{\Sigma}$ can be decomposed into a rotation matrix $\mathbf{R}\in\mathbb{R}^{3\times3}$ and a diagonal scaling matrix $\mathbf{S}\in\mathbb{R}^{3\times3}$ as shown by
\begin{equation}
    \mathbf{\Sigma} = \mathbf{R} \mathbf{S} \mathbf{S}^T \mathbf{R}^T.
\end{equation}
To render novel views, splatting is utilized to project 3D Gaussians onto 2D canvas.
This technique involves a viewing transformation denoted by $\mathbf{W} \in \mathbb{R}^{3\times 3}$ and Jacobian $\mathbf{J} \in \mathbb{R}^{2\times 3}$ of the affine approximation of the projective transformation. 
The 2D covariance matrix $\hat{\mathbf{\Sigma}} \in \mathbb{R}^{2\times 2}$ can then be given as
\begin{equation}
\hat{\mathbf{\Sigma}} = \mathbf{J} \mathbf{W} \mathbf{\Sigma} \mathbf{W}^T \mathbf{J}^T.
\end{equation}
We finally leverage $\alpha$-blending of $N$ overlapped Gaussians at a pixel to accumulate color by
\begin{equation}
\mathbf{c} = \sum_{i = 1} ^ {N} \mathbf{c}_i \alpha_i \prod_{j=1}^{i-1}(1 -\alpha_j),
\label{vr}
\end{equation}
where $\mathbf{c}\in\mathbb{R}^3$ is the rendered pixel color, $\mathbf{c}_i\in\mathbb{R}^3$ and $\alpha_i\in\mathbb{R}$ are color and density of the $i$-th Gaussian point, respectively.

\section{Our Method}
Multi-StyleGS consists of two stages: the reconstruction stage, where a base GS model is trained to recover the original scene and additionally learn semantic correspondences,
and the stylization stage, where the GS model is further refined to adjust its appearance to multiple styles specified by the correspondences.

\subsection{Gaussian Splatting with Semantic Features}
To establish local region correspondences for local style transfer, we leverage segmentation maps and match local regions in 3D scenes with those in style images. In particular, we enhance GS by incorporating an extra segmentation branch, as illustrated in Figure \ref{fig:style_stage}. In addition to the existing attributes of the Gaussians (\textit{e.g.}, color and opacity), we introduce a new trainable feature $\mathbf{e}_i$ for each Gaussian \cite{gaussiangrouping, feature3dgs}. 
This feature $\mathbf{e}_i$ is subsequently decoded by a tiny MLP to predict a semantic category.

To optimize the feature $\mathbf{e}_i$ and the tiny MLP, we render these semantic features into 2D images in a differentiable manner. 
Specifically, we have the following formula for feature integration:
\begin{equation}
    \mathbf{e} = \sum_{i =1}^{N}\mathbf{e}_i \alpha_i \prod_{j=1}^{i-1}(1 -\alpha_j),
\end{equation}
where $\mathbf{e} \in E$ is the rendered feature and $E$ denotes the rendered feature map. The feature $\mathbf{e}$ subsequently passes through a softmax function to calculate the cross-entropy loss $\mathcal{L}^{\textrm{seg}}$. 
However, to compute $\mathcal{L}^{\textrm{seg}}$, we need ground-truth semantic labels. We use SAM~\cite{SAM} to automatically generate semantic labels for each 2D image and employ a well-trained zero-shot tracker~\cite{tracking} to propagate and associate semantic labels~\cite{gaussiangrouping}.  
Back-propagation is employed to optimize the feature $\mathbf{e}_i$ and parameters of the tiny MLP. 

However, we observe that updating each Gaussian point individually can lead to noisy and unstable outcomes due to the stochastic optimization nature and the restricted granularity of points.
To address the issues, we leverage a locality assumption: neighboring points should exhibit similar characteristics. We introduce a regularization loss to enhance the smoothness of segmentation results based on $k$-nearest neighbor (KNN) by
\begin{align}
    \mathcal{L}^{\textrm{KNN}} &= 
    \sum_{i}
    \sum_{j \in \mathcal{N}_i} 
    \Vert \mathbf{e}_i - \mathbf{e}_j \Vert _2^2 \;
    e^{-\frac{\Vert \boldsymbol{\mu}_i - \boldsymbol{\mu}_j \Vert}{\sigma}}  , 
\end{align}
where $\mathcal{N}_i$ gathers $k$ nearest neighbors for the $i$-th Gaussian, and $\sigma \in \mathbb{R}_{>0}$ determines the influence radius. The essence of $\mathcal{L}^{\textrm{KNN}}$ lies in weighting the similarity differences according to the influence of their distances. Our assumption encourages local smoothness, avoids excessive randomness, and increases the granularity of influence.

Moreover, we notice that one Gaussian point may be responsible for multiple objects, leading to semantic ambiguity which is unwanted. We incorporate a negative entropy regularization term:
\begin{align}
    \mathcal{L}_{\textrm{NE}} &= -\sum_{i=1}^{N} \textrm{softmax}(\mathbf{e}_i) \log (\textrm{softmax}(\mathbf{e}_i)),
\end{align}
to enforce each point to choose only one category, eliminating such an ambiguity. 
%

However, some points may be of less semantic importance.
We utilize a semantic importance filter to eliminate those with less semantic significance. Specifically, we additionally assign a learnable mask attribute $m \in \mathbb{R}$ to individual Gaussian~\cite{compact3DGS} to assess its importance, and utilize semantic labels $\textbf{e}_i$ to select Gaussians without semantic ambiguity. We also employ the straight-through estimator~\cite{straight-through} for gradient propagation.
We apply a mask $m^{\textrm{b}} \in \{0, 1\}$ to the scale vector $\mathbf{s} \in \mathbb{R}^{3}$ (diagonal elements of $\mathbf{S}$) and the opacity $o \in \mathbb{R}$ by $\hat{\mathbf{s}} = m^{\textrm{b}} \mathbf{s}$ and $\hat{o} = m^{\textrm{b}} o$, respectively, where the binary mask $m^{\textrm{b}}$ can be obtained by 
\begin{align}
     m^{\textrm{b}} &= \textrm{sg}
    \left[ v^{\textrm{b}} - \sigma\left(m\right)
    \right] + \sigma(m) , \\
    \textrm{with} \quad v^{\textrm{b}} &= \mathds{1}
    _{\sigma \left(m\right) > \epsilon_0 } 
    \lor 
    \mathds{1}
    _{
    \max\left(\textrm{softmax}\left(\mathbf{e}_i\right)\right) > \epsilon_1 
    } 
    \nonumber
    ,
\end{align}
where $\epsilon_0 \in \mathbb{R}$ and $ \epsilon_1 \in \mathbb{R}$ are thresholds, ``$\textrm{sg}$'' is to stop gradients, $\sigma$ is the sigmoid function, $\mathds{1}_{A}$ is an indicator of event $A$, and $\lor$ is logical OR operator.
During reconstruction, the GS model optimizes the scale, opacity, and mask attributes simultaneously. This approach enables a more holistic consideration of both scale and opacity when assessing the importance of Gaussian components.
To promote the decimation of redundant Gaussians, we introduce a mask regularization term given by
\begin{equation}
    \mathcal{L}^{\textrm{mask}} = \sum_{m} \sigma(m)
    .
\end{equation}
We note that by incorporating  $\mathcal{L}^{\textrm{mask}}$, our model facilitates the automatic elimination of Gaussians through gradient control.
By adjusting the weighting coefficient of $\mathcal{L}^{\textrm{mask}}$, we can achieve a more optimal balance between rendering quality and memory footprint.
At specific iterations, we remove certain unnecessary Gaussians based on $m^\textrm{b}$.

\subsection{Preliminary of Style Loss}
Given a pair of rendered output image $I$ and style image $S$, the style loss typically operates on high-level features $\mathbf{f}_I = F(I)$, $\mathbf{f}_S = F(S)$, where $F$ is a pretrained VGG19~\cite{VGG19} network. For instance, StyleGaussian \cite{stylegaussian} employs AdaIN for stylistic transformation, while ARF \cite{ARF} introduces a nearest-neighbor feature matching (NNFM) loss to achieve style transfer. The NNFM loss introduced in \cite{ARF} uses the following formulation,
\begin{align}
    \mathcal{L}_{\textrm{NNFM}}^{\textrm{naive}} = 
    \sum_{f_i \in \mathbf{f}_I} \min_{f_j \in \mathbf{f}_S} d(f_i, f_j)
    ,
\end{align}
where every individual feature vector $f_i \in \mathbf{f}_I$ is paired with the closest style feature $f_j \in \mathbf{f}_S$ according to cosine distance $d$.
However, the style loss is to match the global statistics between the rendered output image $I$ and style image $S$, and can not support diverse stylization results. We incorporate bipartite matching to augment NNFM loss to support local style transfer, which will be detailed in the next coming section.

However, VGG feature is 2D local and has no 3D awareness. Using VGG features only for matching can lead to multi-view inconsistency issues. Moreover, a substantial increase in memory usage is observed when utilizing GS as the base representation for stylization. To address these issues, we propose a novel semantic style loss.

\begin{figure}[t]
\centering
\includegraphics[width=\columnwidth]{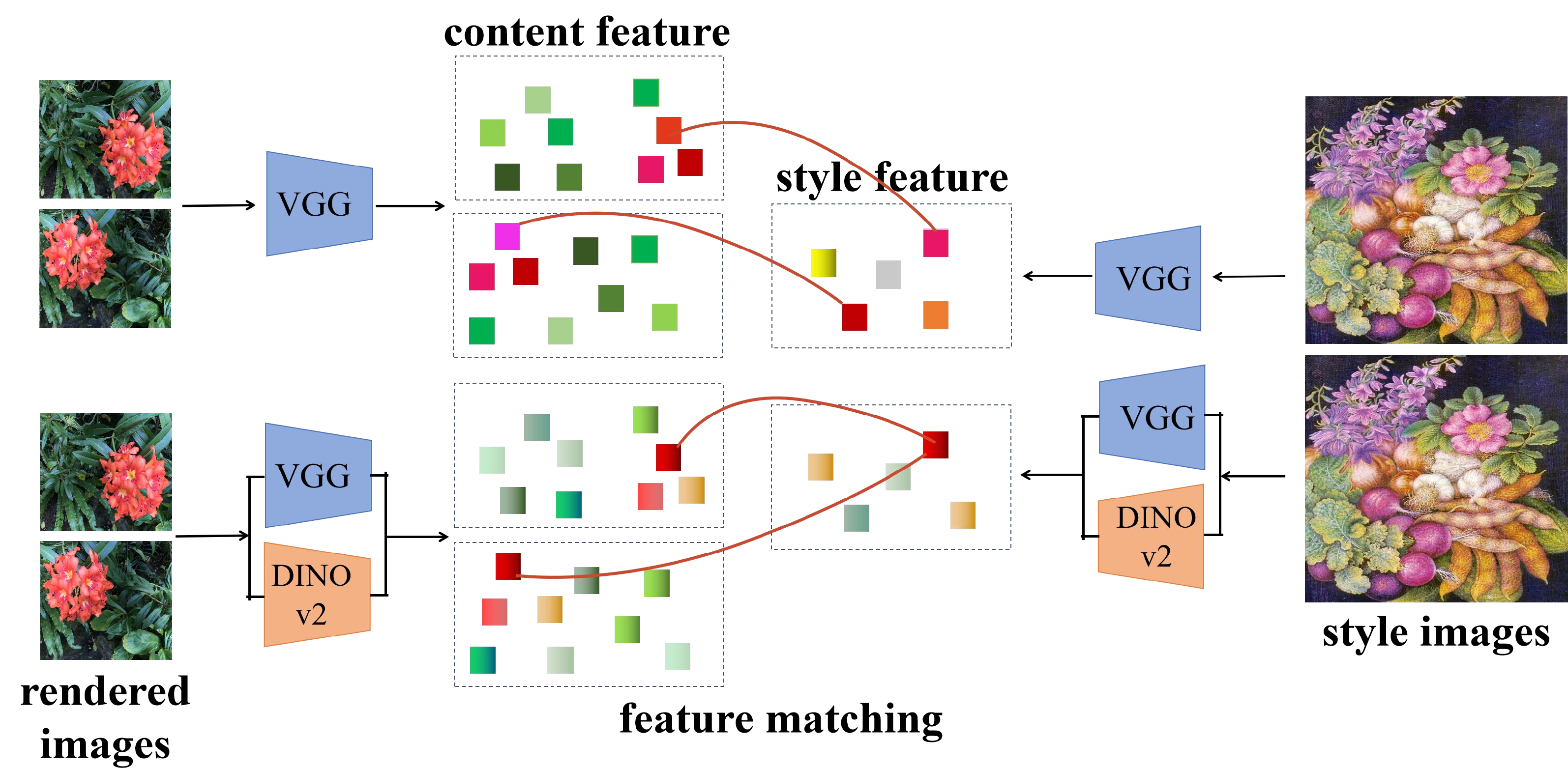} 
\caption{The VGG features do not ensure consistency across different viewpoints, leading to the same object being associated with distinct features when viewed from various angles. However, by incorporating DINOv2, we can maintain local details while also achieving enhanced consistency in feature matching, regardless of the viewing perspective. \vspace{-0.5cm}}
\label{local-global}
\end{figure}
\begin{figure*}[tb]
  \centering
\includegraphics[width=1.0\textwidth]{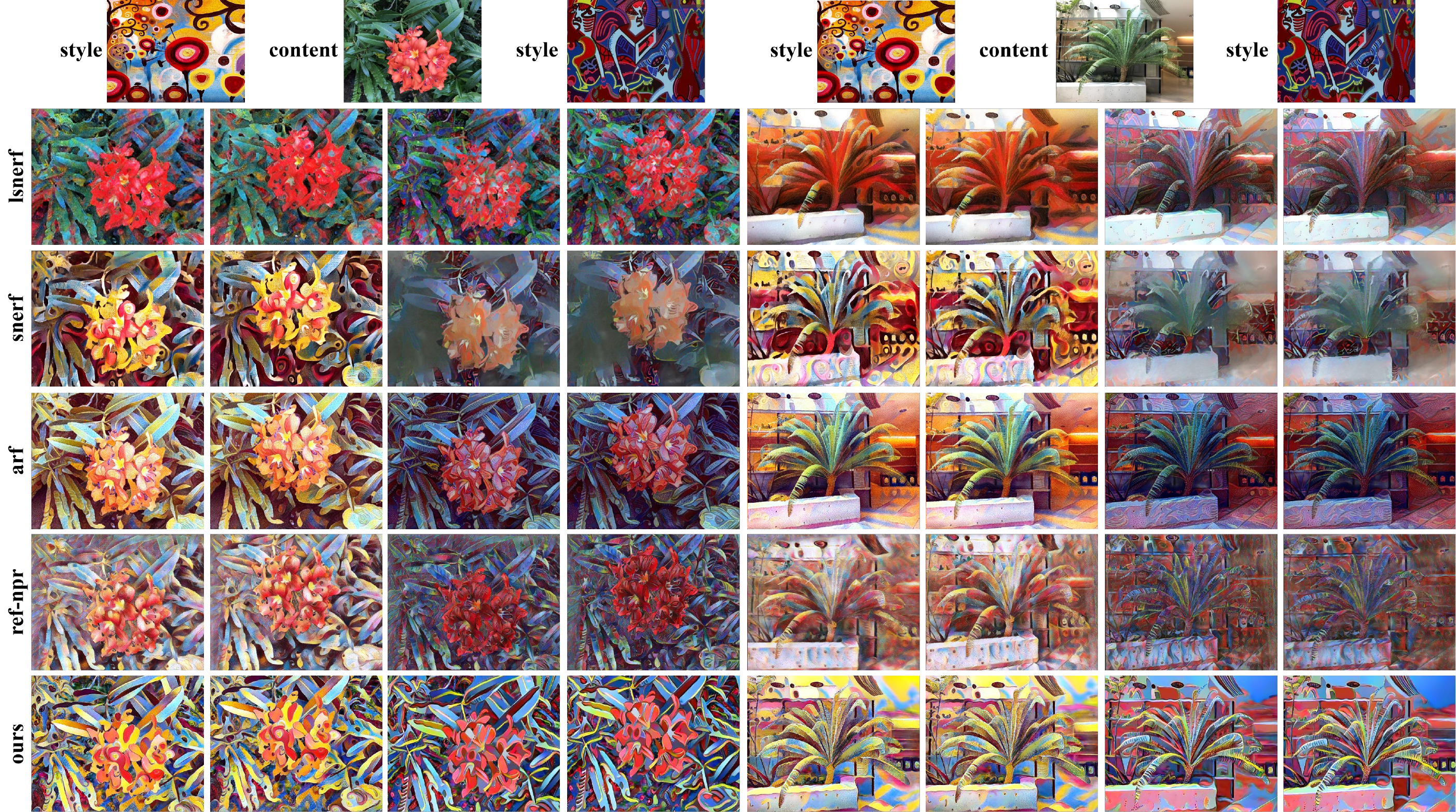}
  \caption{
  Qualitative comparisons with snerf (Nguyen et al. 2022), arf \cite{ARF}, ref-npr \cite{ref-npr} and lsnerf \cite{lsnerf} on \textit{flower} and \textit{fern} scenes \cite{LLFF}.
  \vspace{-0.5cm}
  }
  \label{fig:single_llff}
\end{figure*}
\subsection{Semantic Multi-style Loss}
To facilitate local style transfer, we firstly establish region correspondences for the Gaussian point set $\{g_i\}_{i=1}^N$ of the scene and the set of input style images $\{S_i\}_{i=1}^M$. After reconstruction, the feature $\mathbf{e}_i$ of each Gaussian will indicate the semantic label to which the object it belongs, categorizing the Gaussians into $C$ distinct classes. The initial step in our pipeline involves partitioning $\{g_i\}_{i=1}^N$ into multiple point set $\{G_i\}_{i=1}^C$ as show in Figure \ref{fig:style_stage}.

\subsubsection{Local-global Feature Matching}
Since Gaussian points, when observed from various perspectives, may align with distinct style features, resulting in multi-view inconsistency, as illustrated in Figure \ref{local-global}. We found that features from VGG tend to suffer from such a problem stemming from poor global consistency. 

\begin{figure}[t]
\centering
\includegraphics[width=\columnwidth]{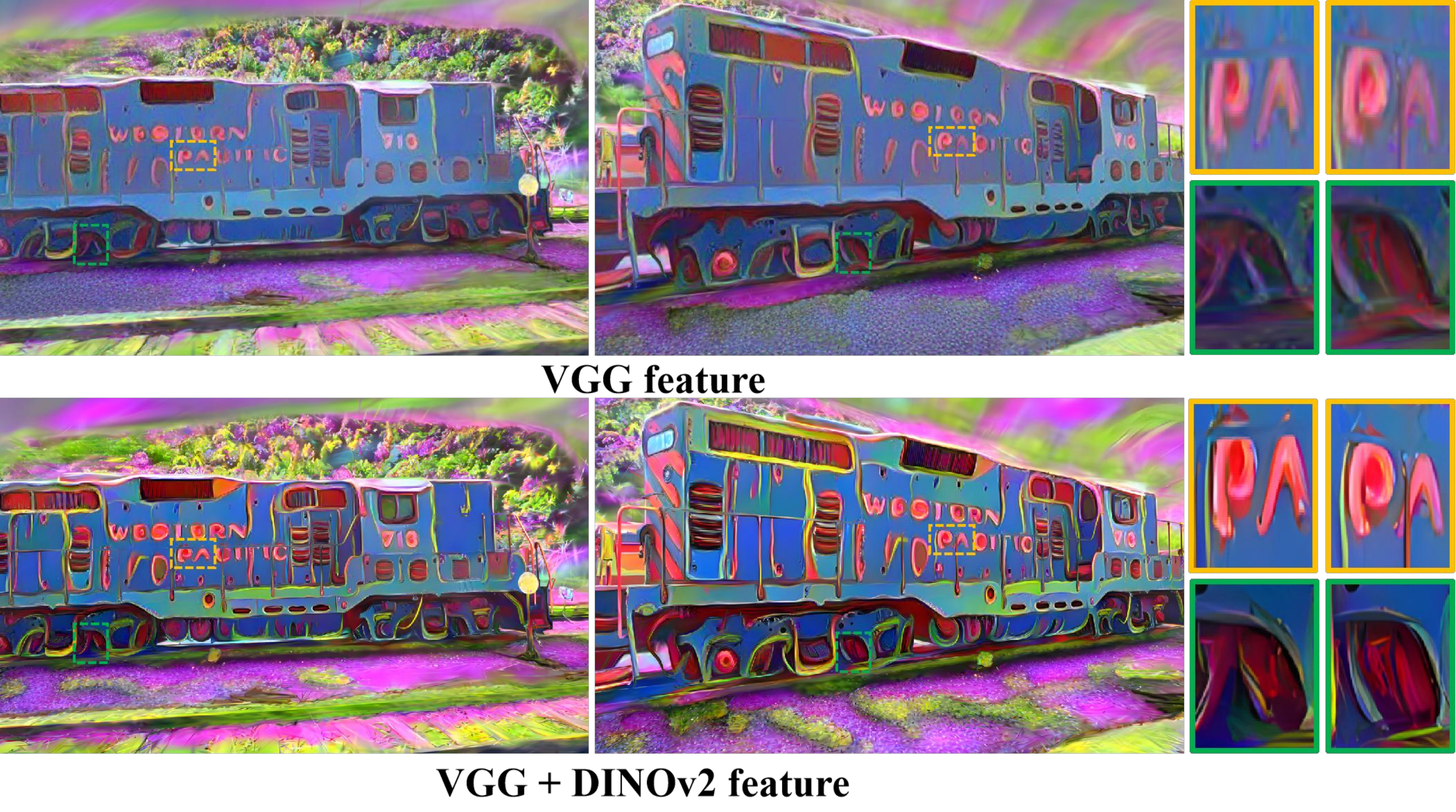} 
\caption{Comparing the stylized results, from various perspectives, the same object may correspond to different VGG features, averaging out the details (as seen in subfigures with orange borders) or displaying varying colors (as seen in subfigures with green borders). DINOv2 enhances the global consistency that VGG features lack, ensuring consistent guidance across different viewpoints.
\vspace{-0.5cm}
}
\label{vgg-dinov2-comparision}
\end{figure}

One paper \cite{Probing3D} assessed the 3D awareness of visual models and posits that DINOv2~\cite{dinov2} demonstrates superior 3D consistency. Therefore, we extract DINOv2 feature and VGG feature and concatenate them along the channel dimension, then perform nearest feature matching on concatenative feature as follows,
\begin{align}
    \mathbf{C}_S = \textrm{concat}(\mathbf{f}_S, \phi(S)),
    \mathbf{C}_I = \textrm{concat}(\mathbf{f}_I, \phi(I)), \\
    \mathcal{L}_{\textrm{NNFM}} = 
    \sum_{f_i \in \mathbf{C}_I} \min_{f_j \in \mathbf{C}_S} d(f_i, f_j)
    ,
\end{align}
where $\phi$ is the DINOv2 feature extractor, ``$\textrm{concat}$'' is to concatenate two feature maps along the channel dimension. VGG feature can provide local details and DINOv2 feature can provide global consistency. $\mathcal{L}_{\textrm{NNFM}}$ not only enhances multi-view consistency but also better improves matching results, ensuring that the same area, when viewed from another perspective, exhibits consistency and is endowed with richer and clearer details, as shown in Figure. \ref{vgg-dinov2-comparision}.

\begin{figure*}[htb]
  \centering
  \includegraphics[width=1.0\textwidth]{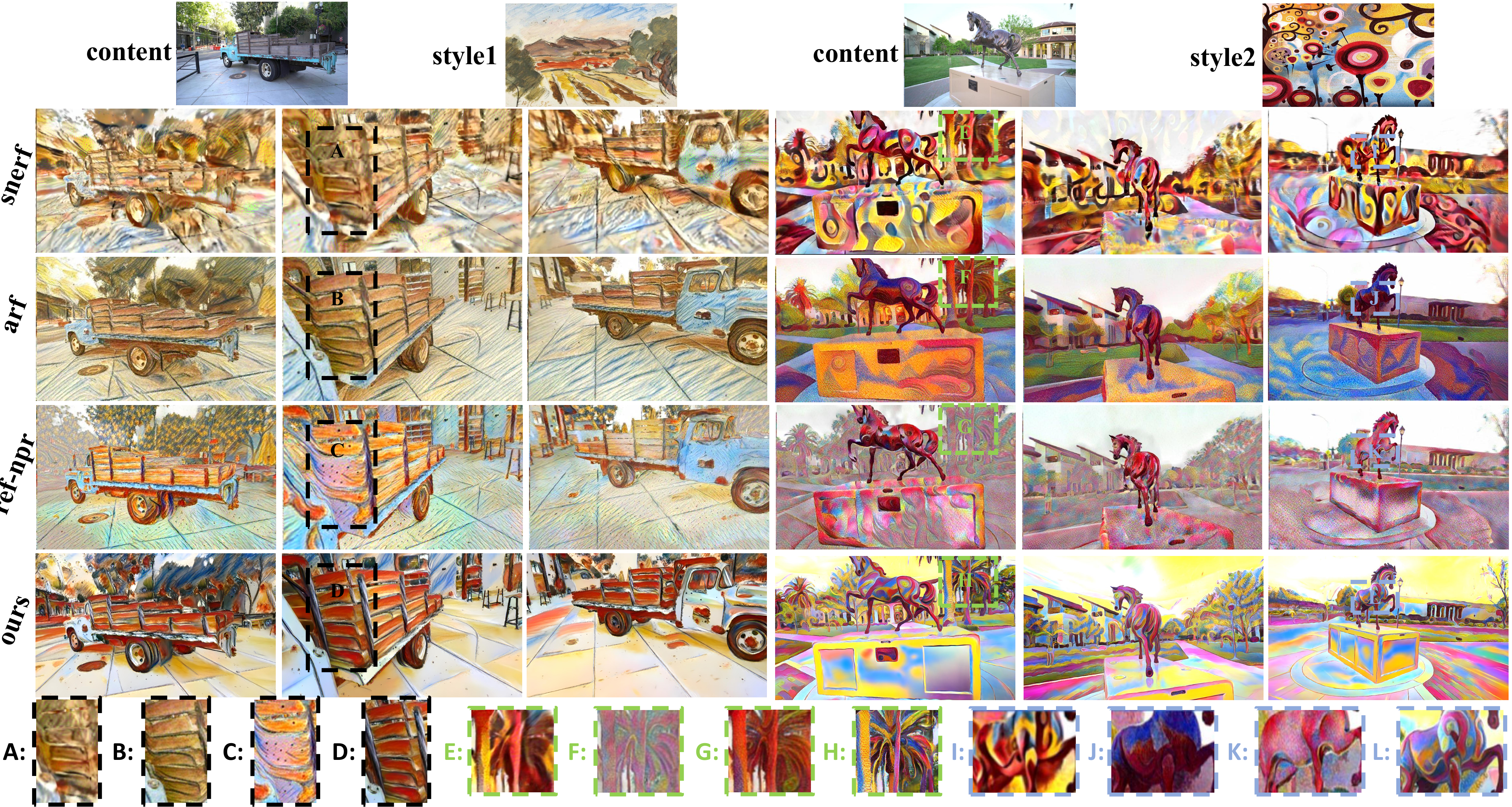}
  \caption{
  Qualitative comparisons with snerf(Nguyen et al. 2022), arf\cite{ARF} and ref-npr\cite{ref-npr} on tnt datasets in single style setting. \textbf{Black boxes (A-D): inconsistency may lead to blurry results}; \textbf{green boxes (E-H): previous solutions may produce incomplete stylized results}; \textbf{blue boxes (I-L): our method can preserve texture details}. Our solution is effective in addressing all of these problems.
    \vspace{-0.5cm}
  }
  \label{fig:single_multi_tnt}
\end{figure*}

\subsubsection{Local Style Loss}
To prevent multiple scene regions from being stylized with the same local pattern, we incorporate a bipartite matching mechanism~\cite{lsnerf} to automatically identify local region correspondences between multiple point set $\{G_i\}_{i=1}^C$ and multiple style images $\{S_i\}_{i=1}^M$.
We construct a cost matrix $\mathbf{Q} \in \mathbb{R}^{C \times M}$, where each entry $Q_{ij}$ represents the correlation between regions $G_i$ and $S_j$. We first render each point set $G_i$ into image $I_{i}$, then utilize VGG19~\cite{CNN19} to extract features from both the rendered image $I_{i}$ and the stylized image $S_j$. The correlation is determined by the cosine feature distance between the means of features of $I_{i}$ and $S_j$.

Given the cost matrix $\mathbf{Q}$, an optimal mapping $\mathcal{M}: [1, C] \mapsto [1, M]$ can be generated by Hungarian algorithm. Our multi-style loss can be finally formulated as 
\begin{equation}
    \mathcal{L}^{\textrm{style}} = 
    \sum_{j=1}^{C}
    \sum_{\mathbf{f} \in C_{I_j}} \min_{\substack{k=\mathcal{M}(j) \\ \mathbf{g} \in C_{S_k}}} d(\mathbf{f}, \mathbf{g}) ,
    \label{localstyleloss}
\end{equation}
where $\mathbf{f}$ and $\mathbf{g}$ are pixel-wise features, $d$ measures the cosine distance. 
Through the minimization of our multi-style loss, we augment the GS model with the ability to perform stylization with multiple styles. Such a design not only enables local style transfer but also significantly alleviates the burden on GPU memory. By strategically categorizing Gaussians into several distinct categories, our model circumvents the need to apply splatting to all Gaussians in a single pass. Moreover, the stylization process with these categorized Gaussians naturally ensures that the resulting appearance exhibits seamless continuity and unambiguity.

\subsection{Training Details}
\subsubsection{Reconstruction stage.} Our GS model is trained with
\begin{equation}
    \mathcal{L}^{\textrm{recon}} + 
    \lambda^{\textrm{seg}} \mathcal{L}^{\textrm{seg}} + 
    \lambda^{\textrm{KNN}} \mathcal{L}^{\textrm{KNN}} +
    \lambda^{\textrm{NE}} \mathcal{L}^{\textrm{NE}}+
    \lambda^{\textrm{mask}} \mathcal{L}^{\textrm{mask}}
    ,
\end{equation}
where $\mathcal{L}^{\textrm{recon}}$ is the Mean Squared Error (MSE) reconstruction loss as outlined in \cite{gs}. We typically assign values of $\lambda^{\textrm{seg}} = 0.02$, $\lambda^{\textrm{KNN}} = 0.005$ , $\lambda^{\textrm{NE}} = 0.005$. 
\subsubsection{Stylization stage.} 
After reconstruction, we can obtain a region mapping $\mathcal{M}$.
During the stylization stage, we utilize the mapping $\mathcal{M}$ and train the model by minimizing
\begin{equation}
    \lambda^{\textrm{cont}} \mathcal{L}^{\textrm{cont}}
     + 
    \lambda^{\textrm{style}} \mathcal{L}^{\textrm{style}}
,
\end{equation}
where $\mathcal{L}^{\textrm{cont}}$ is the content loss, which measures the MSE between the encoded feature map and the ground truth.

\begin{figure*}[htb]
  \centering
  \includegraphics[width=1.0\textwidth]{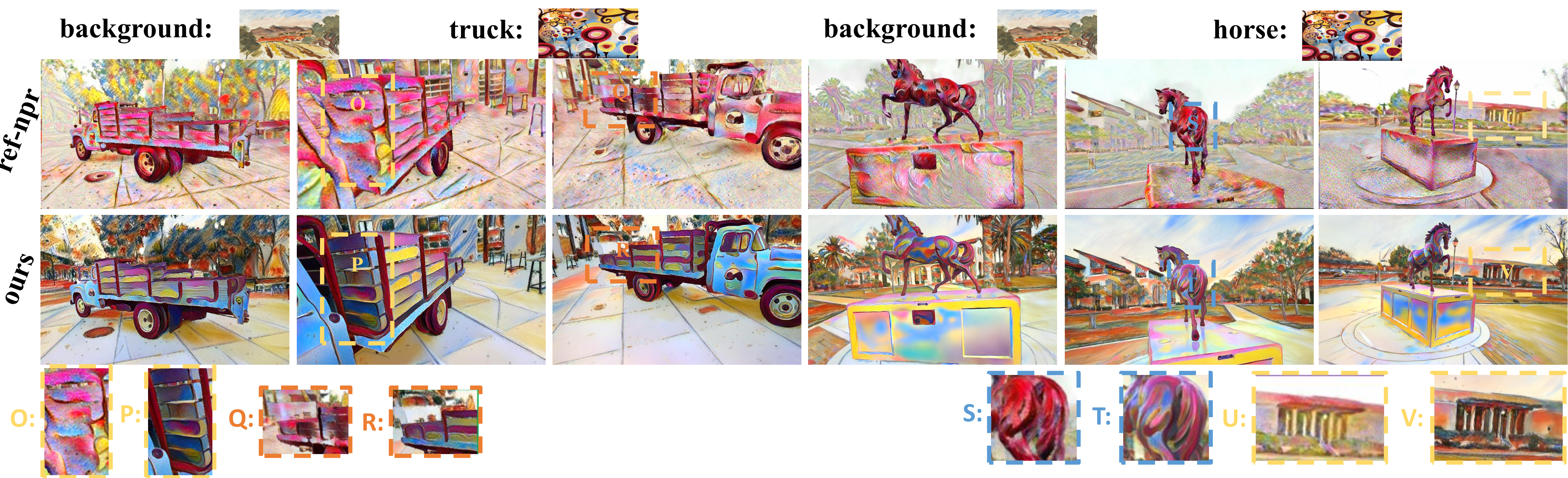}
  \caption{
  Qualitative comparisons with snerf (Nguyen et al. 2022), arf \cite{ARF} and ref-npr \cite{ref-npr} on tnt datasets in multiple style setting. \textbf{Yellow boxes (O, P, U, V): inconsistency can lead to blurry results; orange boxes (Q, R): incorrect color blending between multiple regions; blue boxes (S, T): our method can produce finer texture details.}
  }
  \label{fig:multi_tnt}
\end{figure*}

\vspace{-0.2cm}
\section{Experiments}
\vspace{-0.2cm}
\subsection{Datasets}
We conducted extensive experiments on a diverse set of real-world scenes, including outdoor environments from the Tanks and Temples (shortened as ``tnt'' in our paper) dataset~\cite{tankandtemples} and forward-facing scenes from the llff dataset~\cite{LLFF}.

\vspace{-0.2cm}
\subsection{Evaluation Metrics}
We use Single Image Frechet Inception Distance (SIFID)~\cite{SIFID} to evaluate the stylization similarity. We perform quantitative comparisions on multi-view consistency \cite{HyperNetwork}. Additionally, we provide visual comparisons and results.

\begin{figure}[t]
\vspace{-0.3cm}
\centering
\includegraphics[width=\columnwidth]{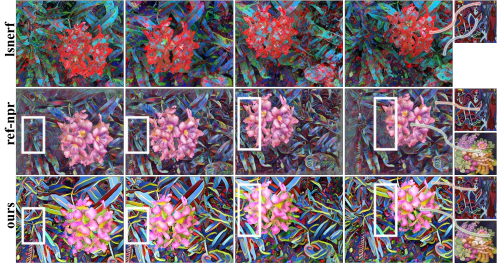} 
\caption{Qualitative comparisions with ref-npr \cite{ref-npr}, lsnerf \cite{lsnerf}. The results enclosed in the white boxes are blurred due to the multi-view inconsistency in the ref-npr. In contrast, our model has effectively ensured consistency.\vspace{-0.5cm}}
\label{local-style}
\end{figure}

\vspace{-0.2cm}
\subsection{Baselines}
On llff datasets~\cite{LLFF}, we compare our method to the SOTA methods, \textit{e.g.}, arf \cite{ARF}, lsnerf \cite{lsnerf}, snerf \cite{snerf} and ref-npr \cite{ref-npr}. Arf and snerf stylize the whole scene with a style image; lsnerf establishs region correspondences between the style image and the content to support local style transfer; ref-npr stylizes scenes with a reference image and VGG matching.

\subsection{Qualitative and Quantitative Comparisons}
In Figure. \ref{fig:single_llff}, we qualitatively compare with other methods on \textbf{llff dataset in single style setting}. Our method provides clearer colors and more accurately stylized texture. Lsnerf fails to fully transfer styles; snerf generates error geometry and produce blurred images; arf employs nnfm in VGG matching, yet suffers from incorrect color blending due to VGG's inability of 3D awareness; ref-npr transfers styles from reference view using VGG but is limited to simple style and struggles with high-frequency signals (\textit{fern} scene). In Figure. \ref{local-style}, we compare our result with two multi-style methods on \textbf{llff datasets in multiply style setting}, lsnerf and ref-npr suffer from multi-view inconsistency, which results in a blurred background. Our method delivers consistent outcomes, detailed textures, and improved color matching.

In Figure. \ref{fig:single_multi_tnt}, we compare our results with four SOTA methods on \textbf{tnt datasets in single style setting}. Snerf generates blurry images due to geometry error; ARF underperforms in the \textit{horse} scene due to its inability to capture fine details and maintain color consistency; ref-npr can only transfer smooth style image and render blurry background in \textit{style2}. Our method adeptly reconstructs scenes, meticulously maintaining the original's geometric and semantic information. Figure. \ref{fig:multi_tnt} presents results with \textbf{multiple styles on tnt datasets}; for the \textit{truck} scene, ref-npr struggles with backgrounds and blends with two styles, whereas our model cleanly distinguishes and separates them from the truck.


\begin{table}[t]
\vspace{-0.3cm}
\hspace{-0.3cm}
\begin{tabular}{ccccc}
\toprule
Methods&truck&horse&flower&Avg. \\
\midrule
snerf (Nguyen et al. 2022) & 3.00 & 2.7 & 0.78 & 2.16 \\
arf \cite{ARF} & 2.23 & 1.50 & 0.16 & 1.29 \\
ref-npr \cite{ref-npr} & 3.12 & 1.78 & 0.88 & 1.92 \\
ours (vgg) & 1.71 & 1.70 & \textbf{0.14} & 1.18 \\
ours & \textbf{1.55} & \textbf{1.21} & \textbf{0.14} & \textbf{0.96} \\
\bottomrule
\end{tabular}
\caption{Quantitative comparisions of multi-view consistency. Lower scores indicate better consistency.\vspace{-0.3cm}}
\label{consistency}
\end{table}


In Table. \ref{consistency}, we use the metrics from \cite{HyperNetwork} to measure the consistency. We generate rendered videos for each scene and randomly sample 50 frames 5 times to calculate their consistency. Our method achieves the best multi-view consistency scores in all metrics.


\vspace{-0.15cm}
\subsection{Ablation Studies}
\vspace{-0.1cm}
\textit{\textbf{Due to limited space, we move additional experimental analysis and results to the supplementary material.}}
\vspace{-0.1cm}
\subsubsection{Semantic Multi-style loss}
Semantic multi-style loss can achieve multi-style transfer and efficient training. Table. \ref{ablation_number} shows the average number of points per iteration when using the multi-style loss to reduce GPU memory usage, \textit{e.g.}, the truck scene that has 1M points in total, is optimized on its categorized subsets typically with 398K points for ``truck'' and 123K points for ``ground''.
\begin{table}[t]
\centering
\begin{tabular}{cccc}
\toprule
style loss & truck & horse & train \\
\midrule
multi-style loss & 362,548.6 & 301,322,7 & 378,561.5\\
nnfm loss & 1,087,646 & 834,546 & 756,747 \\
\bottomrule
\end{tabular}
\caption{Ablation Study of GPU memory. Multi-style loss optimizes fewer points each time, using less GPU memory. \vspace{-0.3cm}}
\label{ablation_number}
\end{table}

\subsubsection{VGG feature and DINOv2 feature}
Figure. \ref{vgg-dinov2-comparision} and Table. \ref{consistency} present the ablation study results of multi-view consistency.
By incorporating DINOv2 features, we can achieve better outcomes with multi-view consistency.

\vspace{-0.3cm}
\section{Limitations and Future Works}

We also notice that our model is incapable of instant style transfer and thus requires retraining for different styles. We will leave this for future improvement.

\bibliography{aaai25}

\newpage

{\large \noindent \textbf{Reproducibility Checklist}}

\vspace{0.2cm}
\noindent This paper:

\begin{itemize}
    \item Includes a conceptual outline and/or pseudocode description of AI methods introduced (yes/partial/no/NA): \textbf{no.}
    \item Clearly delineates statements that are opinions, hypothesis, and speculation from objective facts and results (yes/no): \textbf{yes.}
    \item Provides well marked pedagogical references for less-familiare readers to gain background necessary to replicate the paper (yes/no): \textbf{yes.}
\end{itemize}

\vspace{0.2cm}
\noindent Does this paper make theoretical contributions? (yes/no)  \textbf{no.}
\noindent If yes, please complete the list below.

Does this paper rely on one or more datasets? (yes/no) \textbf{yes}.
\begin{itemize}
    \item A motivation is given for why the experiments are conducted on the selected datasets (yes/partial/no/NA) \textbf{no}.
    \item All novel datasets introduced in this paper are included in a data appendix. (yes/partial/no/NA) \textbf{yes}.
    \item All novel datasets introduced in this paper will be made publicly available upon publication of the paper with a license that allows free usage for research purposes. (yes/partial/no/NA) \textbf{yes}.
    \item All datasets drawn from the existing literature (potentially including authors’ own previously published work) are accompanied by appropriate citations. (yes/no/NA) \textbf{yes}.
    \item All datasets drawn from the existing literature (potentially including authors’ own previously published work) are publicly available. (yes/partial/no/NA) \textbf{yes}.\textbf{NA}
\end{itemize}

Does this paper include computational experiments? (yes/no) \textbf{yes}.
\begin{itemize}
    \item Any code required for pre-processing data is included in the appendix. (yes/partial/no). \textbf{no}.
    \item All source code required for conducting and analyzing the experiments is included in a code appendix. (yes/partial/no). \textbf{no}.
    \item All source code required for conducting and analyzing the experiments will be made publicly available upon publication of the paper with a license that allows free usage for research purposes. (yes/partial/no) \textbf{yes}.
    \item All source code implementing new methods have comments detailing the implementation, with references to the paper where each step comes from (yes/partial/no) \textbf{yes}.
    \item If an algorithm depends on randomness, then the method used for setting seeds is described in a way sufficient to allow replication of results. (yes/partial/no/NA) \textbf{NA}
    \item This paper specifies the computing infrastructure used for running experiments (hardware and software), including GPU/CPU models; amount of memory; operating system; names and versions of relevant software libraries and frameworks. (yes/partial/no) \textbf{no}.
    \item This paper formally describes evaluation metrics used and explains the motivation for choosing these metrics. (yes/partial/no) \textbf{no}.
    \item This paper states the number of algorithm runs used to compute each reported result. (yes/no) \textbf{no}.
    \item Analysis of experiments goes beyond single-dimensional summaries of performance (e.g., average; median) to include measures of variation, confidence, or other distributional information. (yes/no) \textbf{no}.
    \item The significance of any improvement or decrease in performance is judged using appropriate statistical tests (e.g., Wilcoxon signed-rank). (yes/partial/no) \textbf{yes}.
    \item This paper lists all final (hyper-)parameters used for each model/algorithm in the paper’s experiments. (yes/partial/no/NA) \textbf{yes}.
    \item This paper states the number and range of values tried per (hyper-) parameter during development of the paper, along with the criterion used for selecting the final parameter setting. (yes/partial/no/NA) \textbf{NA}
\end{itemize}

\section{Methods, Datasets and Experimental Details}
\subsection{Methods}
We compare with four SOTA methods:
\begin{itemize}
    \item \textbf{codebase:} We are conducting experiments on lsnerf\cite{lsnerf} using its official implementation, and on arf\cite{ARF}, ref-npr\cite{ref-npr}, and snerf\cite{snerf} using the official implementation of ref-npr\cite{ref-npr}, which includes the implementations of these three methods.
    \item \textbf{lsnerf}\cite{lsnerf}: lsnerf is a nerf-based method. lsnerf first trains a segmentation network for 3D scene and use SAM\cite{SAM} to generate segmentation mask for style image, and then generate a regions correspondence between 3D scene and style image. Thus lsnerf can perform local style transfer. 
    Three drawbacks for this methods:
    \begin{enumerate}
        \item Their semantic segmentation only uses cross-entropy loss, so it requires accurate ground truth masks. When the ground truth is not that precise, correct segmentation cannot be achieved. 
        \item In general, different regions of the style image are of the same style (the same material and the same texture), but with different colors. Therefore, in most cases, the results generated by lsnerf can only exhibit color diversity and cannot exhibit diversity in texture and material.
        \item Technical problems: we conduce experiments with lsnerf offical codebase. It can only work on llff\cite{LLFF} datasets. We try replica\cite{replica} and tnt\cite{tankandtemples} datasets and it doesn't work.
    \end{enumerate}

    \item \textbf{snerf}\cite{snerf}: snerf proposes a novel training scheme to reduce GPU usage and several loss to transfer style. However,
    \begin{enumerate}
        \item snerf is capable of learning to transfer the color from a style image, as it relies on statistical methods of image analysis to calculate the loss, which may limit its ability to capture more complex stylistic elements(texture) beyond color.
        \item snerf refines the geometry but lacks of efficient supervision. In some cases, blurring may occur due to geometric errors.
    \end{enumerate}

    \item \textbf{arf}\cite{ARF}: arf proposes nnfm loss to perform precise texture transfer, and deferred back-propagation for memory reduction. However,
    \begin{enumerate}
        \item nnfm loss uses the VGG\cite{VGG19} features of the style and rendering for matching. The VGG features of the same object in different view may match different style feature due to the VGG feature lacks of 3D awareness, lead to the same object displaying different styles from different views.
        \item When rendering the full-resolution image, the deferred back-propagation method still consumes a significant amount of GPU memory.
    \end{enumerate}

    \item \textbf{ref-npr}\cite{ref-npr}: ref-npr first styles a reference view, then transfer the style of reference view to other views. We use SANET\cite{SANET} to style reference view with style image. ref-npr introduces the Template-Based Semantic Correspondence (TCM) mechanism, leveraging VGG features for nearest neighbor searches, facilitating the seamless diffusion of stylistic elements from the reference view to alternative perspectives. However,
    \begin{enumerate}
        \item VGG features have no 3D awareness and using VGG for nearest search across different view can lead to multi-view inconsistency issues.
        \item If the stylied reference image contain complex texture details, pixels from certain perspectives may not match to the pixels in the reference view, hence style transfer cannot be performed, resulting in white noise spots.
    \end{enumerate}
\end{itemize}

\begin{figure*}[htb]
  \centering
  \includegraphics[width=1.0\textwidth]{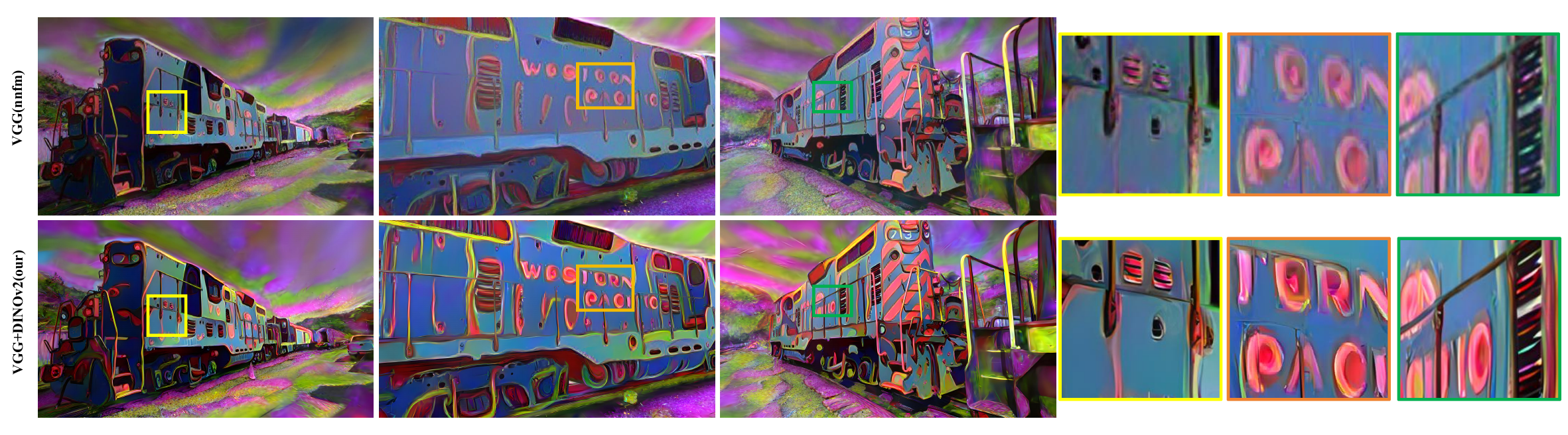}
  \caption{
  Ablation study of texture. Second row is multi-style loss with local-global matching, first row is nnfm loss\cite{ARF} with VGG feature matching.
  }
  \label{ablation_texture_v2}
\end{figure*}

\begin{figure}[t]
\centering
\includegraphics[width=\columnwidth]{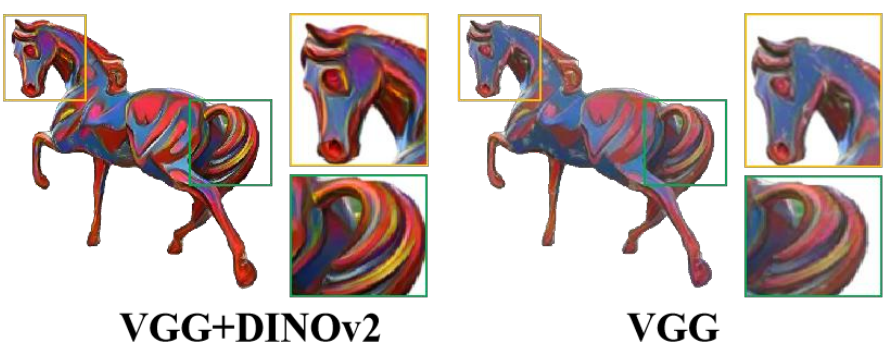} 
\caption{Ablation Study of texture. Left image is multi-style loss with local-global matching, right image is nnfm loss\cite{ARF} with VGG feature matching.
}
\label{ablation_texture}
\end{figure}

\subsubsection{Discussion of Controllable Style Methods}
There are some controllable style transfer methods: StyleRF\cite{styleRF}, ConRF\cite{ConRF}, ref-npr\cite{ref-npr}, lsnerf\cite{lsnerf}. StyleRF require segmentation mask for all training images. For large scenes containing hundreds of images, it is impractical to generate masks for all images, and multi-view consistency cannot be guaranteed; ConRF use Clip\cite{CLIP} to generate segmentation mask for each training image and multi-view consistency cannot be guaranteed; lsnerf trains a segmentation network, but they can only perform segmentation in relatively simple scenes and require accuracy ground-truth masks; ref-npr can control the style of the scene by adjusting the style of the reference view. 

StyleRF and ConRF are zero-shot style transfer methods, and ConRF can not handle 360 degree scene(tnt\cite{tankandtemples}) datasets. Thus we did not compare with StyleRF and ConRF.

\subsection{Datasets}
We conduce experiments on two datasets: llff\cite{LLFF} datasets and tnt\cite{tankandtemples} datasets.
\begin{itemize}
    \item \textbf{llff}\cite{LLFF}: simple dataset only contains mimic variation of view angles. llff dataset is to easy for Gaussian Splatting\cite{gs}, thus we conduct experiments on more complex 3D scenes\cite{tankandtemples}.
    \item \textbf{tnt}\cite{tankandtemples}: 360◦ captures outdoor scenes.
\end{itemize}

\begin{figure}[t]
\centering
\includegraphics[width=\columnwidth]{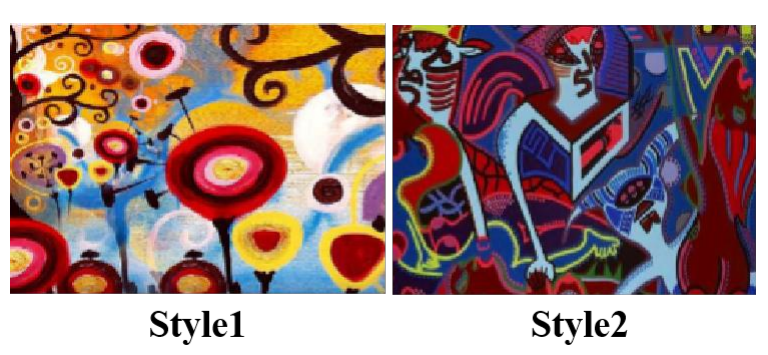} 
\caption{Figure. 4 style images
}
\label{llff_style_images}
\end{figure}

\begin{figure*}[htb]
  \centering
  \includegraphics[width=1.0\textwidth]{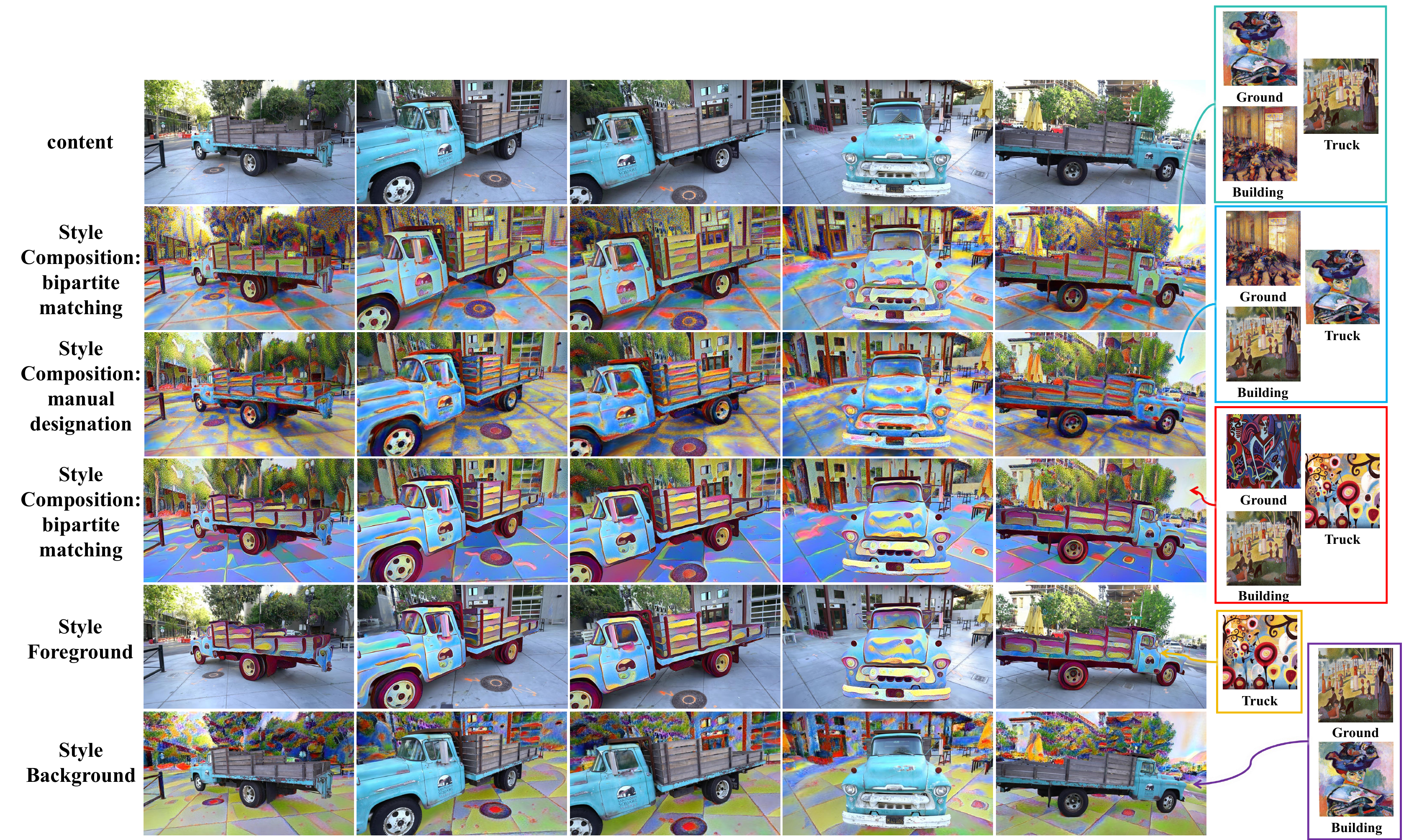}
  \caption{
  Ablation Study of multi-style transfer of Multi-Style loss. 
  \textbf{Second row and third row}: ablation study of the bipartite matching. The style matching can be assigned through an bipartite matching or manual designation; \textbf{Second row and fourth row:} ablation study of the diverse style. We can set arbitrary styles for different areas. \textbf{Last two row:} ablation study of on a single object. We can specify any object for arbitrary styling.
  }
  \label{fig:ablation_style}
\end{figure*}

\subsection{Experiments Explaination}
\begin{figure}[t]
\centering
\includegraphics[width=\columnwidth]{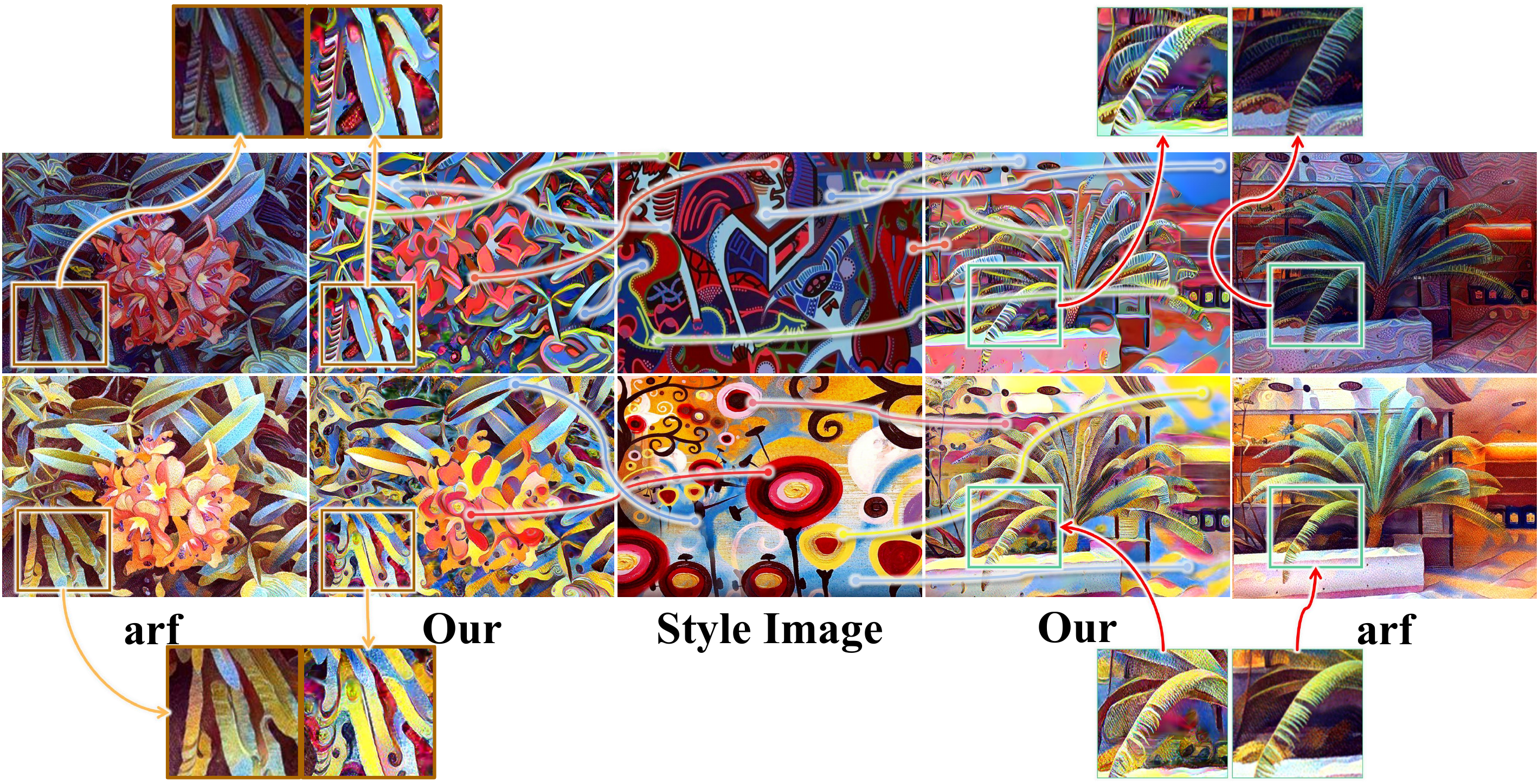} 
\caption{The colors present in the style image can almost all be found in the images we generate. In contrast, the results from arf\cite{ARF} cannot find the green color in \textit{flower} scene in the first row, and arf generates image with darker hues by blending color. Additionally, our method contain more texture details.}
\label{color_match}
\end{figure}

Experiments Explaination for Figure. 4, 8, 6, 7 in regular paper.
\begin{itemize}
    \item Figure. 4: single style transfer on llff datasets with two style images shown on Figure. \ref{llff_style_images}.
        \subitem lsnerf failed to stylize in \textit{flower} scene because the matching mechanism matched different regions in the \textit{flower} scene to a small regions of the style image, which led to the failure in capturing the local style of the style image. In \textit{fern} scene, the images generated by LSNeRF exhibit a monotonous color palette and instances of blurriness, such as in the rendering of leaves.
        \subitem The primary issues with snerf arise from geometric error that lead to multi-view inconsistencies and image blurriness. 
        \subitem arf can generate the desired results in \textit{Style1}. In \textit{Style2},  arf blended all the colors together to render a stylized image with darker hues, resulting in a color mismatch issue.
        \subitem ref-npr can produce appropriate stylization in \textit{Style1}, but when faced with a more complex stylization image like \textit{Style2}, it tends to generate images that are blurry and unclear. Because relying solely on VGG cannot ensure the consistency of matches across different views. This issue becomes more severe in the \textit{train} scene in tnt\cite{tankandtemples} datasets in Figure. \ref{fig:additional_train_single_0}.
        \subitem As shown in Figure. \ref{color_match}, our method can generate more stable results. First, the colors present in the style image can almost all be found in the images our mothed generate. In contrast, in the \textit{fern} scene on \textit{Style2}, the images generated by ARF lack green color. Second, our method can generate more texture details. Third, our method can perform multi-style transfer.

        \item Figure. 8: multiple styles transfer on llff datasets. lsnerf can only perform local style transfer on a single image, whereas ref-npr can perform multi-style transfer on the reference view and then proceed with 3D style transfer. ref-npr may exhibit blurriness at the boundary of different styles because it does not use semantics to accurately segment different objects.

        \item Figure. 6 and Figure. 7: style transfer on tnt datasets. Compare with these methods, 
            \subitem firstly, our method can generate complete style transfer(Figure 6, the fourth row, ref-npr  has overlooked the background);
            \subitem Secondly, our method can achieve correct color transfer(Figure 6, the third row, \textit{horse} scene, arf generates colors are too dark and do not match the colors of the style image.)
            \subitem Thirdly, our method can generate more precise texture. And in multiple styles in Figure. 7, our model can ensure that the styles between two different objects do not interfere with each other(Figure 7, the second row, orange box, The styles generated by ref-npr of the foreground and background are mixed together, with the background presents red color).
\end{itemize}

\section{Additional Experiments}
We primarily conducted ablation studies on the style loss and three regularization terms.
\subsection{Ablation Study}
\subsubsection{Multi-Style Loss}
\begin{figure}[t]
\centering
\includegraphics[width=\columnwidth]{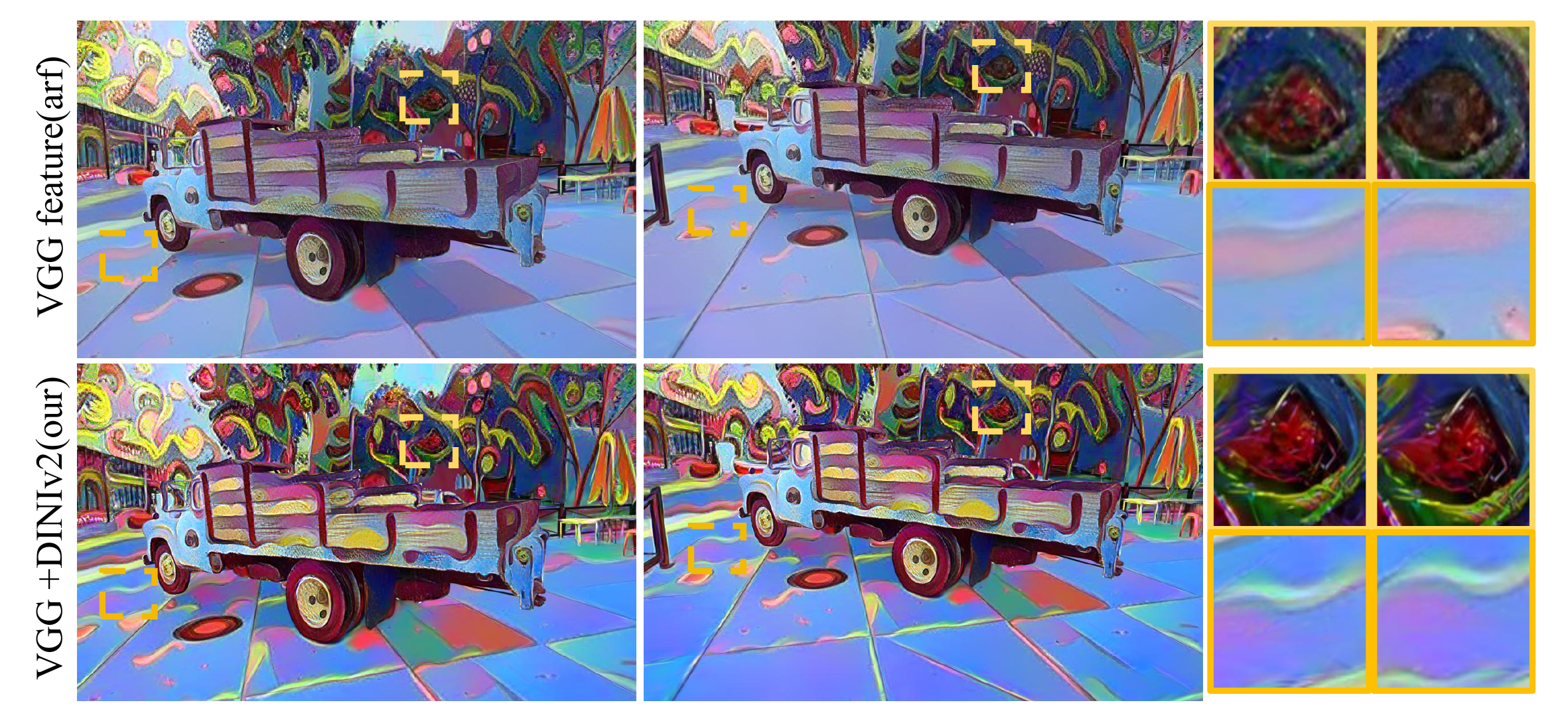} 
\caption{Ablation study of consistency, in the \textit{truck} scene, from various perspectives, the same object may correspond to different VGG features, averaging out the details (as seen in subfigures with orange borders) or displaying varying colors (as seen in subfigures with green borders). DINOv2 enhances the global consistency that VGG features lack, ensuring consistent guidance across different viewpoints.}
\label{ablation_consistency}
\end{figure}

\begin{figure}[t]
\centering
\includegraphics[width=\columnwidth]{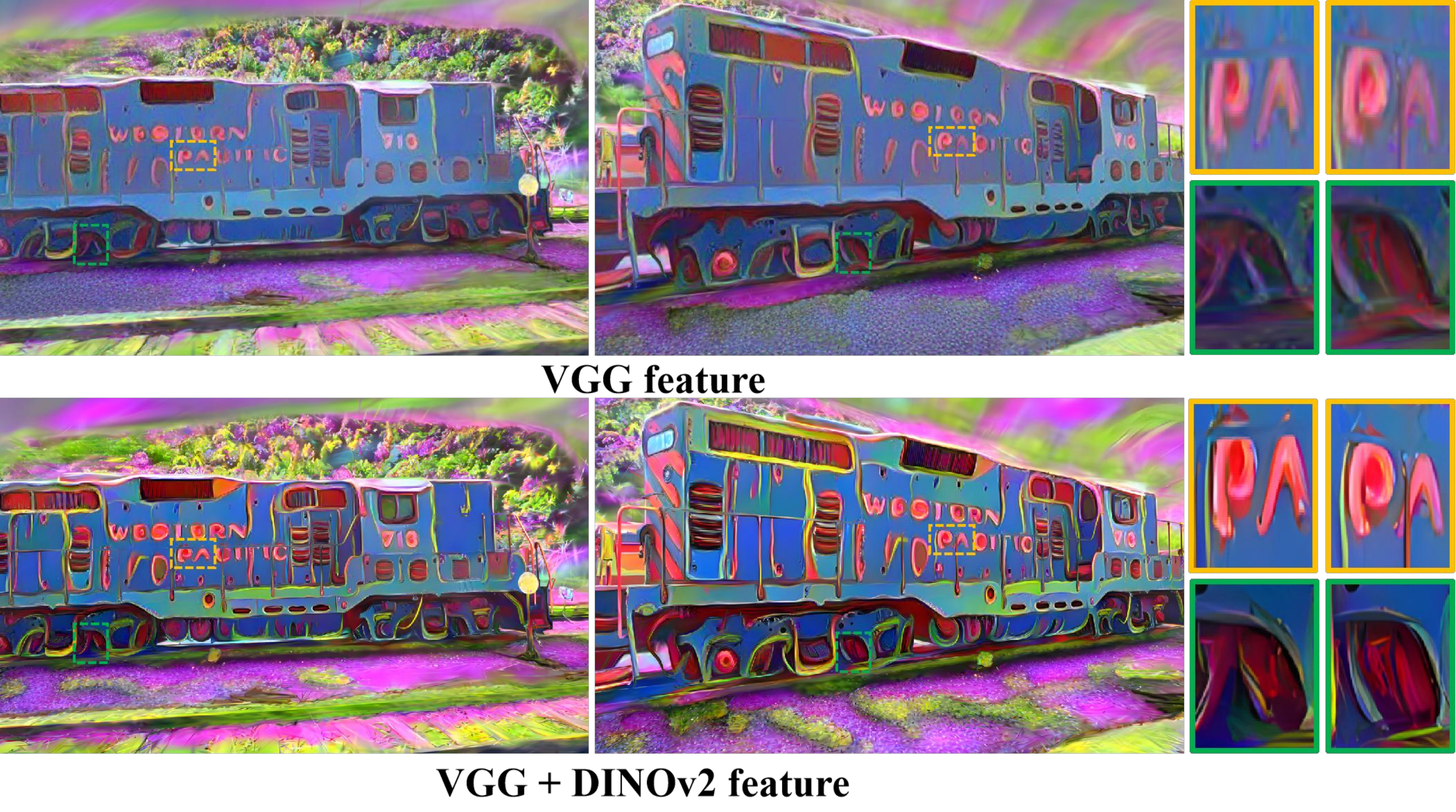} 
\caption{Ablation study of consistency, in the \textit{train} scene,from various perspectives, the same object may correspond to different VGG features, averaging out the details (as seen in subfigures with orange borders) or displaying varying colors (as seen in subfigures with green borders). DINOv2 enhances the global consistency that VGG features lack, ensuring consistent guidance across different viewpoints.}
\label{vgg_dino_comparision}
\end{figure}

Three advances for multi-style loss: 
(1) perform multi-style transfer; (2) reduce GPU memory usage; (3) richer texture; (4) multi-view consistency. 
\begin{itemize}
    \item Multi-Style Transfer: We segment the scene into multiple parts, each part styles transfer with a different style. As shown in Figure. \ref{fig:ablation_style}, our multi-style loss can perform multiple style, single and stylization of a single object.
    \item GPU Memory Usage: Because we have divided the scene into different parts, we can optimize a selection of Gaussian points for each part, reducing memory usage. Table. \ref{ablation_number} shows the number of Gaussian points that optimized in each iteration.
    \item Richer Texture: Figure. \ref{ablation_texture} and Figure. \ref{ablation_texture_v2} demonstrates the effectiveness of our method, capable of learning clearer and richer textural details.
    \item Multi-view Consistency: 
        \subitem As shown in Figure. \ref{ablation_consistency} and Figure. \ref{vgg_dino_comparision}, in terms of visual quality, our method can ensure multi-view consistency across different views. 
        \subitem Additionally, we have provided a consistency Table. \ref{consistency} for all methods. (1) We use the metrics from \cite{HyperNetwork} to measure the consistency. (2) For a fair comparison, all methods are only stylized in a single style. (3) And we randomly sampled 15 images from both the testing trajectory and the training trajectory, making a total of 30 images for consistency evaluation. ref-npr exhibits poor consistency on the \textit{train} scene, which is why we did not test it on the \textit{train} scene. Our method can achieve the best consistency score in all metrics.
\end{itemize}

\begin{table}[t]
\vspace{-0.3cm}
\hspace{-0.3cm}
\begin{tabular}{cccccc}
\toprule
Methods&truck&horse& Family &flower&Avg. \\
\midrule
snerf & 3.00 & 2.7 & 2.9 & 0.78 & 2.35 \\
arf & 2.23 & 1.50 & 1.43 & 0.16 & 1.33 \\
ref-npr & 3.12 & 1.78 & 1.83 & 0.88 & 1.90 \\
ours (vgg) & 1.71 & 1.70 & 1.53 & \textbf{0.14} & 1.27 \\
ours & \textbf{1.55} & \textbf{1.21} & \textbf{1.35} & \textbf{0.14} & \textbf{1.06} \\
\bottomrule
\end{tabular}
\caption{Quantitative comparisions of multi-view consistency. Lower scores indicate better consistency.\vspace{-0.3cm}}
\label{consistency}
\end{table}

\begin{table}[t]
\centering
\begin{tabularx}{0.5\textwidth}{XXXXX}
\toprule
style loss & truck & horse & train & Family \\
\midrule
multi-style loss(our) & 362,548.6 & 301,322,7 & 378,561.5 & 333,453.5\\
nnfm loss(arf) & 1,087,646 & 834,546 & 756,747 & 1,232,545 \\
gram loss(snerf) & 1,087,646 & 834,546 & 756,747 & 1,232,545 \\
\bottomrule
\end{tabularx}
\caption{Ablation Study of GPU memory. Multi-style loss optimizes fewer points each time, using less GPU memory. \vspace{-0.3cm}}
\label{ablation_number}
\end{table}

\begin{figure*}[htb]
  \centering
  \includegraphics[width=1.0\textwidth]{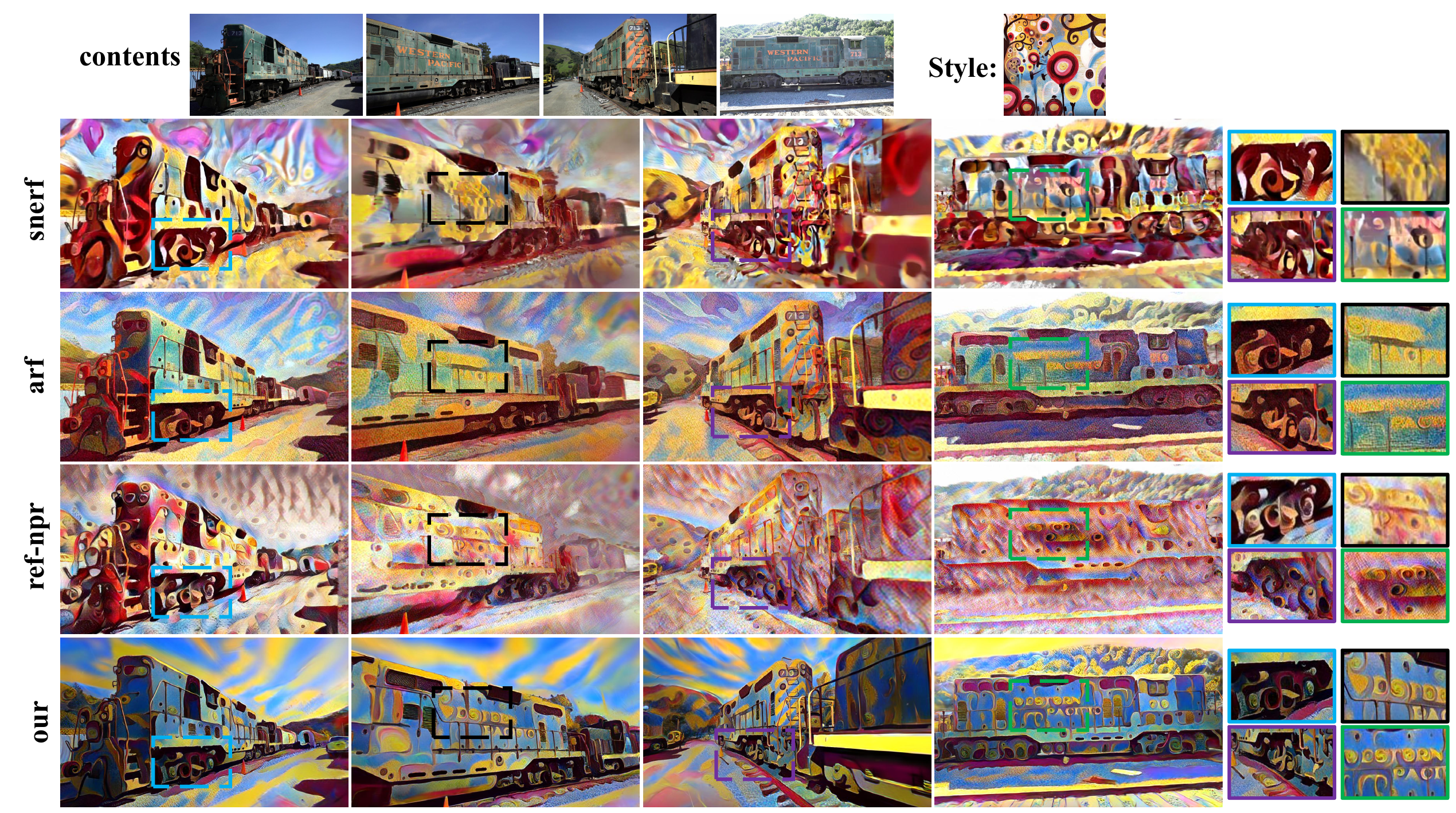}
  \caption{
  Aditional experiments results on \textit{train} scene in single style setting. \textbf{Blue and purple box}: clearer geometry structure; \textbf{Black and green box}: texture and details.
  }
  \label{fig:additional_train_single_0}
\end{figure*}

\begin{figure*}[htb]
  \centering
  \includegraphics[width=1.0\textwidth]{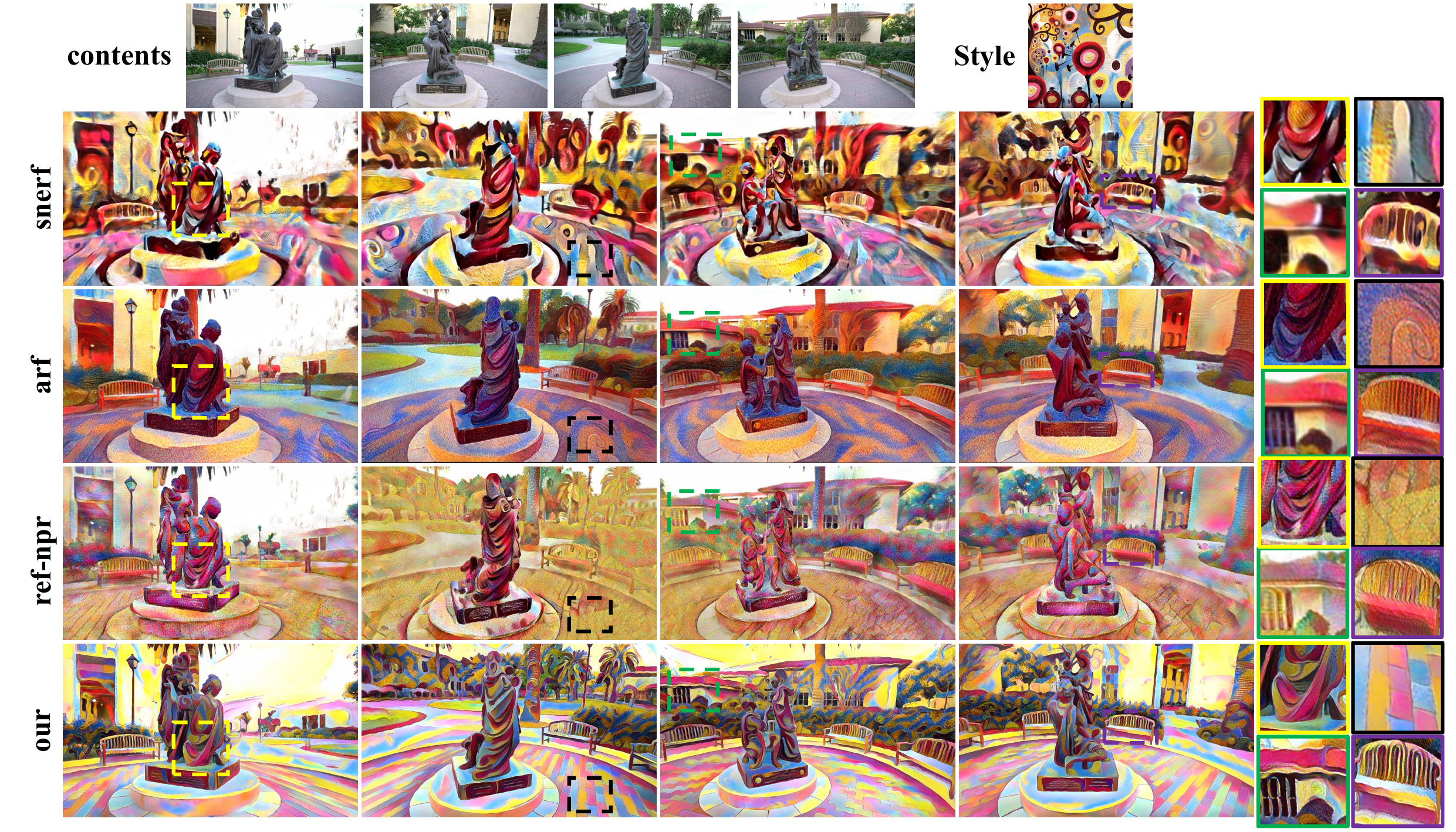}
  \caption{
  Aditional experiments results on \textit{Family} scene in single style setting.
  }
  \label{fig:additional_family_single_0}
\end{figure*}

\subsubsection{Regularizations}
Given that the labels for our semantic segmentation are derived from video segmentation\cite{DEVA}, which do not guarantee high-quality annotations, we have introduced three regularization techniques to significantly improve the accuracy of our semantic segmentation. KNN smooth regularization ensures the correction of some detail errors, while negative entropy and semantic importance filter ensures the correction of large-area errors. \textbf{Negative entropy must be used in conjunction with semantic importance filter; Negative entropy identifies Gaussians in the scene with significant semantic content, while semantic importance filter eliminates the remaining Gaussians that do not convey semantic information. Therefore, using negative entropy alone is ineffective.} We conduct two ablation study: (1) knn regularization; (2)negative entropy and semantic importance filter. As shown in Figure. \ref{ablation_regularization}, our model can ensure the completeness of both segmentation details and the overall integrity.

\begin{figure}[t]
\centering
\includegraphics[width=\columnwidth]{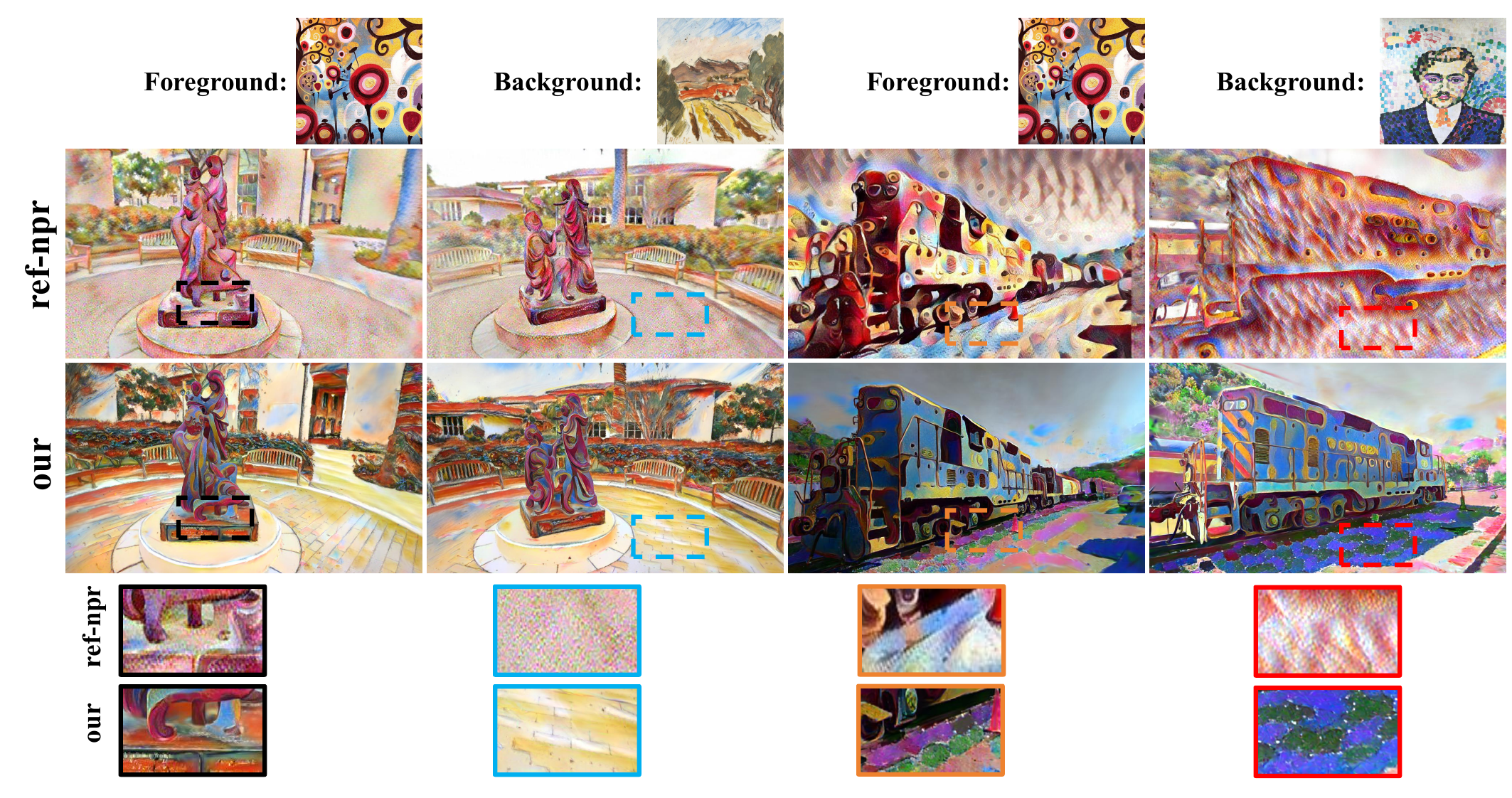} 
\caption{Aditional experiments results on \textit{train} and \textit{Family} scene in multiple style setting. 
}
\label{addition_experiments_multi_style}
\end{figure}

\begin{figure}[t]
\centering
\includegraphics[width=\columnwidth]{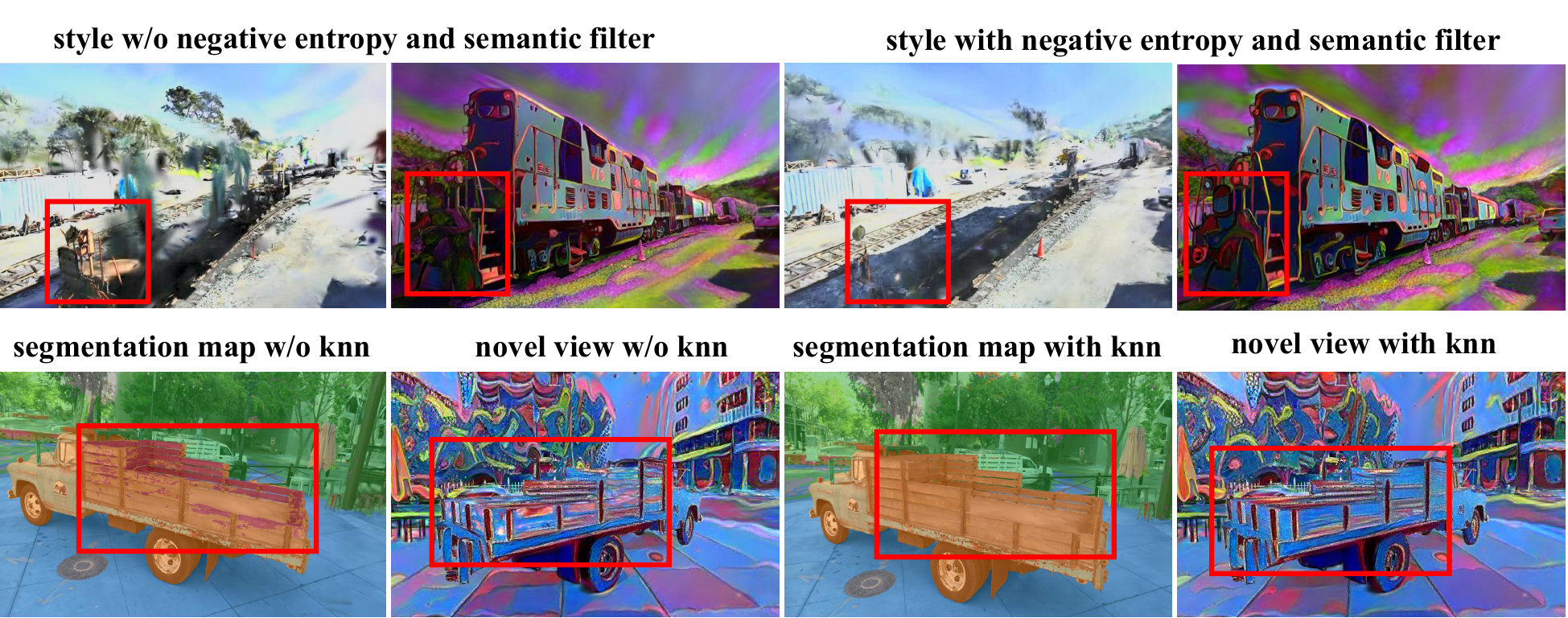} 
\caption{Ablation study of regularization. Our regularization can correctly guide the style, leading to an appropriate stylization.}
\label{ablation_regularization}
\end{figure}

\subsection{Additional Experiments}
We conduct additional experiments on \textit{train} and \textit{Family} scene in tnt\cite{tankandtemples} to evaluate efficiency. Due to the difficulty of the \textit{train} dataset, ref-npr uses 5 styled reference views for stylization.

Figure. \ref{fig:additional_train_single_0} and Figure. \ref{fig:additional_family_single_0} show the single style transfer on \textit{train} and \textit{family} scene. snerf has produced blurry results due to incorrect geometry; arf can generate decent results, but there are issues with texture blurring and unclear structure; ref-npr\cite{ref-npr} leverages VGG features for pixel-level matching across various viewpoints; however, this approach encounters challenges with multi-view consistency. Our model is capable of generating clear and rich details while ensuring multi-view consistency. Additionally, our method can preserve structure information of the original scene(black box in Figure. 0, our method can preserve the texture of the ground).

Figure. \ref{addition_experiments_multi_style} shows the multi-style transfer on \textit{train} and \textit{family} scene. ref-npr is unable to transfer the style of the background and cannot distinguish between two different styles, whereas our method can effectively migrate different styles to their respective objects.

\subsection{Additional Metrics}
 We present the results of a user study designed to assess visual appeal based on user preferences. We collected scores from 27 participants for each set of stylization results produced by ref-npr, arf, and
snerf, and then computed the average scores(scale of ten) for 5 different stylized scenes. In multi-style setting, we randomly combine from 5 different styles. And we conduct user study on \textit{truck}, \textit{horse} and \textit{flower} scenes, in these three scenarios, all comparative methods can generate appropriate stylized results.

As shown in Table. \ref{user_study}, notably, our proposed method outperforms the others, achieving the highest average score. Additionally, when we show our multi-style results, most users have raised their scores.

We provide a table showing the average styling time for four methods, as shown in Table. \ref{time}.

\begin{table}[t]
\centering
\begin{tabularx}{0.5\textwidth}{XX}
\toprule
method & styling time \\
\midrule
Arf & 21.2m \\
SNerf & 184.1m  \\
ref-npr & 16.3m  \\
our(vgg) & 7.1m \\
our(single-style) & 23.8m \\
our(multi-style) & 22.4m \\
\bottomrule
\end{tabularx}
\caption{Ablation style of styling time.}
\label{time}
\end{table}

\begin{table}[t]
\centering
\begin{tabularx}{0.5\textwidth}{XXXXX}
\toprule
score & truck & horse & Flower & avg. \\
\midrule
snerf & 3.4 & 3.3 & 4.0 & 3.56 \\
arf & 7.2 & 8.0 & 5.6 & 6.93 \\
ref-npr & 5.2 & 4.4 & 7.6 & 5.73 \\
our(single) & 8.2 & 7.1 & 7.2 & 7.50 \\
our(all) & \textbf{8.6} & \textbf{7.4} & \textbf{7.3} & \textbf{7.76} \\
\bottomrule
\end{tabularx}
\caption{User Study. \textbf{our(single)}: we only show the single style transfer results; \textbf{our(all)}: we show the single and multiple style transfer results. 
Most users approve of our multi-style stylization, and our scores have improved after demonstrating the multi-stylized results. }
\label{user_study}
\end{table}

\end{document}